\pdfoutput=1 
\documentclass[10pt,letterpaper]{article}
\usepackage[letterpaper,total={6.5in, 9in}]{geometry}
\usepackage[utf8]{inputenc} 
\usepackage[T1]{fontenc}    
\usepackage{hyperref}       
\usepackage{url}            
\usepackage{booktabs}       
\usepackage{amsfonts}       
\usepackage{nicefrac}       
\usepackage{microtype}      
\usepackage{booktabs,caption}
\usepackage[flushleft]{threeparttable}
\usepackage{times}
\usepackage{graphicx} 
\usepackage{enumitem}
\usepackage[numbers,square]{natbib}
\usepackage{helvet}
\usepackage{mathptmx} 
\usepackage{mathpazo} 
\usepackage{titlesec}
\usepackage{authblk}
\usepackage[labelfont=bf,textfont=it,font=footnotesize]{caption}

\usepackage{algorithmic}
\usepackage{xcolor}
\usepackage[linesnumbered,ruled,vlined]{algorithm2e}

\SetCommentSty{mycommfont}

\usepackage{graphicx,color}
\usepackage{epsfig}
\usepackage{amsmath,amsthm,amssymb}
\usepackage{multirow}
\newtheorem{thm}{Theorem}
\newtheorem{assumption}{Assumption}

\newtheorem{lem}{Lemma}
\newtheorem{proposition}{Proposition}
\newtheorem{defn}{Definition}

\theoremstyle{remark}
\newtheorem{rem}{Remark}

\newcommand{\N}{\mathbb{N}}

\newcommand{\x}{\mathbf{x}}

\newcommand{\E}{\mathbb{E}}
\graphicspath{{figures/}}

\usepackage{subcaption}

\titleformat{\section}{\large\bfseries\sffamily}{\thesection}{1em}{}
\titleformat{\subsection}{\normalsize\bfseries\sffamily}{\thesubsection}{1em}{}
\titleformat{\subsubsection}{\small\sffamily\bfseries}{\thesubsubsection}{1em}{}

\usepackage{hyperref}
\hypersetup{
	colorlinks,
	linkcolor={magenta},
	citecolor={blue},
	urlcolor={blue!80!black},
	plainpages=true
}
\newcommand{\cref}[2]{\hyperref[#2]{#1~\ref*{#2}}}

\begin{document}

\title{\usefont{OT1}{phv}{b}{}\selectfont\LARGE{On Consensus-Optimality Trade-offs in Collaborative Deep Learning}}

\author[1]{\usefont{OT1}{phv}{}{}\selectfont\large
	Zhanhong Jiang}
\author[1]{Aditya Balu}
\author[2]{Chinmay Hegde}
\author[1]{Soumik Sarkar}
\affil[1]{\usefont{OT1}{phv}{}{}\selectfont\large
	Department of Mechanical Engineering, Iowa State University}
\affil[2]{\usefont{OT1}{phv}{}{}\selectfont\large
	Department of Electrical and Computer Engineering, Iowa State University\\
	\tt\small {\{zhjiang|baditya|chinmay|soumiks\}}@iastate.edu}
\date{}
\maketitle

\begin{abstract}
In distributed machine learning, where agents collaboratively learn from diverse private data sets, there is a fundamental tension between \emph{consensus} and \emph{optimality}. 
In this paper, we build on recent algorithmic progresses in distributed deep learning to explore various consensus-optimality trade-offs over a fixed communication topology. First, we propose the~\textit{incremental consensus}-based distributed SGD (i-CDSGD) algorithm, which involves multiple consensus steps (where each agent communicates information with its neighbors) within each SGD iteration. 
Second, we propose the~\textit{generalized consensus}-based distributed SGD (g-CDSGD) algorithm that enables us to navigate the full spectrum from complete consensus (all agents agree) to complete disagreement (each agent converges to individual model parameters). 
We analytically establish convergence of the proposed algorithms for strongly convex and nonconvex objective functions; we also analyze the momentum variants of the algorithms for the strongly convex case. 
We support our algorithms via numerical experiments, and demonstrate significant improvements over existing methods for collaborative deep learning.
\end{abstract}

\section{Introduction}

Scaling up deep learning algorithms in a distributed setting~\cite{lecun2015deep,recht2011hogwild,jin2016scale} is becoming increasingly critical, impacting several applications such as learning in robotic networks~\cite{lenz2015deep}, the Internet of Things (IoT)~\cite{gubbi2013internet,lane2015early}, and mobile device networks~\cite{lane2015can}. Several distributed deep learning approaches have been proposed to address issues such as model parallelism~\cite{dean2012large}, data parallelism~\cite{dean2012large,jiang2017collaborative}, and the role of communication and computation~\cite{li2014communication,das2016distributed}. 

\begin{figure}
\centering
\includegraphics[width=0.4\textwidth, trim={0.5cm, 0.0cm, 0.7cm, 0.0cm},clip]{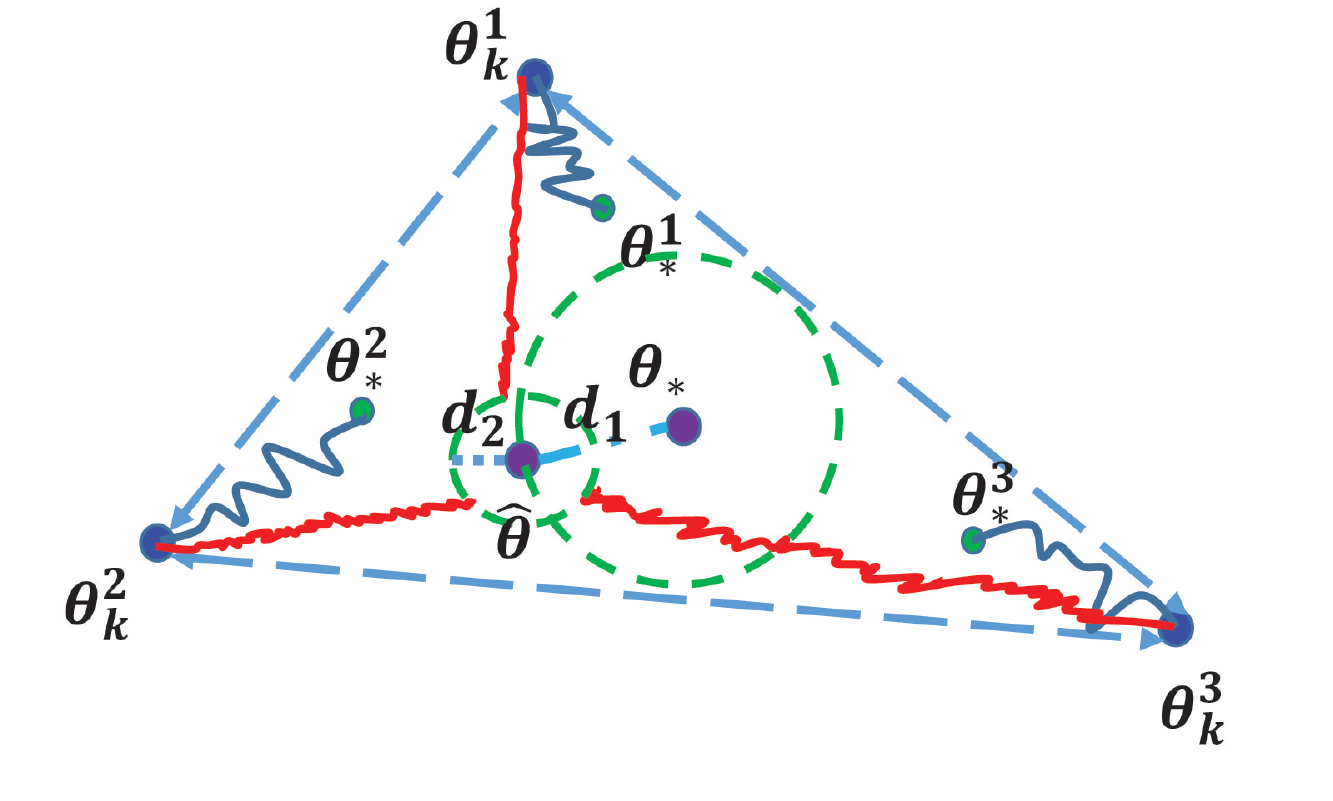}
\caption{\textit{A closer look at the optimization updates in distributed deep learning: Blue dots represent the current states (i.e., learned model parameters) of the agents; green dots represent the individual local optima ($\theta^i_*$), that agents converge to without sufficient consensus; the purple dot ($\theta_*$) represents the ideal optimal point for the entire agent population; another purple dot $\hat{\theta}$ represents a possible consensus point for the agents which is far from optimal; blue and red curves signify the convergence trajectories with different step sizes; the green dashed circles indicate the neighborhoods of $\theta_*$ and $\hat{\theta}$, respectively; $d_2$ represents the consensus bound/error and $d_1$ represents the optimality bound/error; ideally, both of these bounds should be small.}}\label{Figure1}
\end{figure}

We focus on the constrained communication topology setting where the data is distributed (so that each agent has its own estimate of the deep model) and where information exchange among the learning agents are constrained along the edges of a given communication graph~\cite{jiang2017collaborative,lian2017can}. In this context, two key aspects arise: \emph{consensus} and \emph{optimality}. We refer the reader to Figure~\ref{Figure1} for an illustration involving 3 agents. With sufficient information exchange, the learned model parameters corresponding to each agent, $\theta^j_k, j=1,2,3$ could converge to $\hat{\theta}$, in which case they achieve consensus but not optimality (here, $\theta_*$ is the optimal model estimate if all the data were centralized). On the other hand, if no communication happens, the agents may approach their individual model estimates ($\theta^i_*$) while being far from consensus. The question is whether this trade-off between consensus and optimality can be balanced so that \emph{all} agents collectively agree upon a model estimate close to $\theta_*$. 


\textbf{Our contributions}: In this paper, we propose, analyze, and empirically evaluate two new algorithmic frameworks for distributed deep learning that enable us to explore fundamental trade-offs between consensus and optimality. The first approach is called \textit{incremental consensus}-based distributed SGD (i-CDSGD), which is a stochastic extension of the descent-style algorithm proposed in~\cite{berahas2017balancing}. This involves running multiple consensus steps where each agent exchanges information with its neighbors within each SGD iteration. The second approach is called~\textit{generalized consensus}-based distributed SGD (g-CDSGD), based on the concept of generalized gossip~\cite{jiang2017generalised}. This involves a tuning parameter that explicitly controls the trade-off between consensus and optimality.  Specifically, we:
\begin{itemize}[nolistsep,leftmargin=*,topsep=0pt]
\item (\textbf{Algorithmic}) propose the i-CDSGD and g-CDSGD algorithms (along with their momentum variants).
\item (\textbf{Theoretical}) prove the convergence of g-CDSGD (Theorems 1 \& 3) and i-CDSGD (Theorems 2 \& 4) for strongly convex and non-convex objective functions;
\item (\textbf{Theoretical}) prove the convergence of the momentum variants of g-CDSGD (Theorem 5) and i-CDSGD (Theorem 6) for strongly convex objective functions;
\item (\textbf{Practical}) empirically demonstrate that i-CDMSGD (the momentum variant of i-CDSGD) can achieve similar (global) accuracy as the state-of-the-art with lower fluctuation across epochs as well as better consensus;
\item (\textbf{Practical}) empirically demonstrate that g-CDMSGD (the momentum variant of g-CDSGD) can achieve similar (global) accuracy as the state-of-the-art with lower fluctuation, smaller generalization error and better consensus.
\end{itemize}
We use both balanced and unbalanced data sets (i.e., equal or unequal distributions of training samples among the agents) for the numerical experiments with benchmark deep learning data sets. Please see Table~\ref{table1} for a detailed comparison with existing algorithms. 

\begin{table*}
\begin{center}
\resizebox{\textwidth}{!}{%
\begin{threeparttable}
\caption{Comparisons between different optimization approaches}\label{table1}
  \begin{tabular}{c c c c c c c c}
    \hline
    Method & $f$ & Con.Bou. & Opt.Bou. & Con.Rate& Mom.Ana.  & C.C.T. &Sto.\\ \hline
    FedAvg~\cite{mcmahan2016communication} & Nonconvex & N/A & N/A  &N/A&  No & No &Yes\\
    \textnormal{$DGD^\tau$}~\cite{berahas2017balancing} & Str-con & $\mathcal{O}(\frac{\alpha}{1-\lambda_2^\tau})$ &$\mathcal{O}(\frac{\alpha}{1-\lambda_2^\tau})$& $\mathcal{O}(\epsilon^k)$ & No & Yes & No \\ 
    MSDA~\cite{scaman2017optimal}& Str-con&N/A&N/A&$\mathcal{O}(\epsilon^k)$&Yes&Yes&No\\ \cline{2-8}
    \multirow{2}{*}{CDSGD~\cite{jiang2017collaborative}} & Str-con & $\mathcal{O}(\frac{\alpha}{1-\lambda_2})$ &$\mathcal{O}(\frac{\alpha\gamma+1}{H+\alpha^{-1}(1-\lambda_2)})$&$\mathcal{O}(\epsilon^k)$  & \multirow{2}{*}{No} & \multirow{2}{*}{Yes} & \multirow{2}{*}{Yes}\\
    &Nonconvex&$\mathcal{O}(\frac{\alpha}{1-\lambda_2})$&$\mathcal{O}(\alpha \gamma+1-\lambda_N)$&N/A\\ \cline{2-8}
    Acc-DNGD-SC~\cite{qu2017accelerated}&Str-con&$\mathcal{O}(\frac{\alpha^{\frac{1}{3}}}{(1-\lambda_2)\lambda_2^{\frac{2}{3}}})$&N/A&$\mathcal{O}(\epsilon^k)$&Yes&Yes&No\\ \cline{2-8}
    \multirow{2}{*}{i-CDSGD [This paper]}& Str-con & $\mathcal{O}(\frac{\alpha}{1-\lambda_2^\tau})$ &$\mathcal{O}(\frac{\alpha\gamma+1}{H+\alpha^{-1}(1-\lambda_2^\tau)})$& $\mathcal{O}(\epsilon^k)$ & \multirow{2}{*} {Yes} & \multirow{2}{*} {Yes} & \multirow{2}{*} {Yes}\\
    & Nonconvex &$\mathcal{O}(\frac{\alpha}{1-\lambda_2^\tau})$& $\mathcal{O}(\alpha\gamma +1-\lambda_N^\tau)$ & N/A \\ \cline{2-8}
    \multirow{2}{*}{g-CDSGD [This paper]} & Str-con & $\mathcal{O}(\frac{\omega \alpha}{1-\lambda_2})$ &$\mathcal{O}(\frac{\alpha\gamma-1+\omega^{-1}}{H})$& $\mathcal{O}(\epsilon^k)$ & \multirow{2}{*} {Yes} & \multirow{2}{*}{Yes} & \multirow{2}{*}{Yes}\\
    & Nonconvex &$\mathcal{O}(\frac{\omega \alpha}{1-\lambda_2})$ &$\mathcal{O}(\omega\alpha\gamma+(1-\omega)(1-\lambda_N))$& N/A \\
    \hline
  \end{tabular}

\begin{tablenotes}
    \small\sl
    \item Con.Bou.: consensus bound. Opt.Bou.: optimality bound. Con.Rate: convergence rate. Str-con: strongly convex. Mom.Ana.: momentum analysis. $\alpha$: step size. $\lambda_2\in(0,1)$: the second largest eigen-value of a stochastic matrix. $\tau\in\mathbb{N}$: positive constant. $\omega\in(0,1]$: a positive constant. $\epsilon\in (0,1)$: a positive constant, and it signifies the representative meaning. They are not exactly the same in different methods. Sto.: stochastic. C.C.T.: constrained communication topology. $c_1, c_2\textgreater 0$: condition numbers. $H$: strong convexity constant. $\gamma_1, \gamma_2, \gamma_3\textgreater 0$: smoothness constants. 
  \end{tablenotes}
\end{threeparttable}
}
\end{center}
\vspace{-15pt}
\end{table*}

\textbf{Related work}: A large literature has emerged that studies distributed deep learning in both centralized and decentralized settings~\cite{dean2012large,mcmahan2016communication,zhang2015deep,blot2016gossip,jin2016scale,xu2017adaptive,zheng2016asynchronous,zhang2017projection}, and due to space limitations we only summarize the most recent work.~\cite{wangni2017gradient} propose a gradient sparsification approach for communication-efficient distributed learning, while ~\cite{wen2017terngrad} propose the concept of ternary gradients to reduce communication costs.~\cite{scaman2017optimal} propose a multi-step dual accelerated method using a gossip protocol to provide an optimal decentralized optimization algorithm for smooth and strongly convex loss functions. Decentralized parallel stochastic gradient descent~\cite{lian2017can} has also been proposed. 

Perhaps most closely related to this paper is the work of~\cite{berahas2017balancing}, who present a distributed optimization method (called $DGD^\tau$) to enable consensus when the cost of communication is cheap. However, the authors only considered convex optimization problems, and only study deterministic gradient updates. Also,~\cite{qu2017accelerated} propose a class of (deterministic) accelerated distributed Nesterov gradient descent methods to achieve linear convergence rate, for the special case of strongly convex objective functions. 
In~\cite{tsianos2012distributed}, both deterministic and stochastic distributed were discussed while the algorithm had no acceleration techniques. To our knowledge, none of these previous works have explicitly studied the trade-off between consensus and optimality.

\textbf{Outline}: 
Section~\ref{problem_formulation} presents the problem and several mathematical preliminaries. In Section~\ref{proposed_algorithm}, we present our two algorithmic frameworks, along with their analysis in Section~\ref{main_results}. For validating the proposed schemes, several experimental results based on benchmark data sets are presented in Section~\ref{experimental_results}. Concluding remarks are in Section~\ref{conclusion}.

\section{Problem Formulation}
\label{problem_formulation}

We consider the standard unconstrained empirical risk minimization (ERM) problem typically used in machine learning problems (such as deep learning):
\begin{equation}\label{setup}
\text{min} \, \frac{1}{n}\sum_{i=1}^{n}f^i(\theta),
\end{equation}
where $\theta \in \mathbb{R}^d$ denotes the parameter vector of interest, $f:\mathbb{R}^d\to\mathbb{R}$ denotes a given loss function, and $f^i$ is the function value corresponding to a data point $i$. Our focus is to investigate the case where the ERM problem is solved collaboratively among a number of computational \emph{agents}.
In this paper, we are interested in problems where the agents exhibit \emph{data parallelism}, i.e., they only have access to their own respective training datasets. However, we assume that the agents can communicate over a static undirected graph $\mathcal{G} = (\mathcal{V},\mathcal{E})$, where $\mathcal{V}$ is a vertex set (with nodes corresponding to agents) and $\mathcal{E}$ is an edge set.
Throughout this paper we assume that the graph $\mathcal{G}$ is \emph{connected}. 

Let $\mathcal{D}_j,~j=1,\ldots,n$ denote the subset of the training data (comprising $n_j$ samples) corresponding to the $j^{\textrm{th}}$ agent such that $\sum_{j=1}^N n_j=n$, where $N$ is the total number of agents. With this formulation, and since $f(\theta) = \sum_{j=1}^N f_j(\theta)$, we have the following (constrained) reformulation of ~\eqref{setup}:
\begin{equation}\label{d_setup_3}
\text{min} \, \sum_{j=1}^{N}\sum_{i\in \mathcal{D}_j}f^i_j(\theta^j),\;\text{s.t.~} \, \theta^j=\theta^l ~~ \forall (j,l)\in \mathcal{E} ,
\end{equation}
Equivalently, the concatenated form of the above equation is as follows:
\begin{equation}\label{ob_c}
\text{min} \, \mathcal{F}(\Theta):=\sum_{j=1}^{N}\sum_{i\in \mathcal{D}_j}f^i_j(\theta^j),\;\text{s.t.~} \, (\Pi\otimes I_d)\Theta = \Theta,
\end{equation}
where $\Theta := [\theta^1; \theta^2; \ldots; \theta^N]^T\in \mathbb{R}^{dN}$, $\Pi\in\mathbb{R}^{N\times N}$ is the agent interaction matrix with its entries $\pi_{jl}$ indicating the link between agents $j$ and $l$, $I_d$ is the identity matrix of dimension $d\times d$, and $\otimes$ represents the Kronecker product.

We now introduce several key definitions and assumptions that characterize the above problem.
\begin{defn}\label{strong_convexity_smooth}
A function $f:\mathbb{R}^d\to \mathbb{R}$ is said to be $H$-strongly convex, if for all $x,y\in \mathbb{R}^d$, we have $f(y)\geq f(x)+\nabla f(x)^T(y-x)+\frac{H}{2}\|y-x\|^2$;
it is said to be $\gamma$-smooth if we have
$f(y)\leq f(x)+\nabla f(x)^T(y-x)+\frac{\gamma}{2}\|y-x\|^2$; it is said to be $h$-Lipschitz continuous  if we have $|f(y) - f(x)| < h\|y-x\|$. Here, $\|\cdot\|$ represents the Euclidean norm. 
\end{defn}


\begin{defn}\label{coercivity}
A function $c$ is said to be coercive if it satisfies:
$c(x)\to \infty\;when\;\|x\| \to \infty.$
\end{defn}

\begin{assumption}\label{assump_1}
The objective functions $f_j:\mathbb{R}^d\to\mathbb{R}$ are assumed to satisfy the following conditions: a) each $f_j$ is $\gamma_j$-smooth; b) each $f_j$ is proper (not everywhere infinite) and coercive; c) each $f_j$ is $h_j$-Lipschitz continuous.
\end{assumption}


\begin{assumption}\label{assump_2}
The interaction matrix $\Pi$ is normalized to be doubly stochastic; the second largest eigenvalue of $\Pi$ is strictly less than 1, i.e., $\lambda_2(\Pi)\textless 1$, where $\lambda_2(\Pi)$ is the second largest eigenvalue of $\Pi$. If $(j,l)\notin \mathcal{E}$, then $\pi_{jl} = 0$.
\end{assumption}

For convenience, we use $\lambda_2$ to represent $\lambda_2(\Pi)$ and similar $\lambda_N$ for $\lambda_N(\Pi)$, which signifies the $N$-largest eigenvalue of $\Pi$. An immediate consequence from Assumption~\ref{assump_1} (c) is that $\sum_{j=1}^{N}f_j(x)$ is $h$-Lipschitz continuous, where $h = \max\limits_{1\leq j\leq N}\{h_j\}$.

We will solve \eqref{d_setup_3} in a distributed and stochastic manner.
For solving stochastic optimization problems, variants of the well-known stochastic gradient descent (SGD) have been commonly employed. For the formulation in~\eqref{d_setup_3}, the state-of-the-art algorithm is a method called \emph{consensus distributed SGD}, or CDSGD, recently proposed in~\cite{jiang2017collaborative}. This method estimates $\theta$ according to the update equation:
\begin{equation}\label{update_law}
\theta^j_{k+1} = \sum_{l\in Nb(j)}\pi_{jl}\theta^l_k-\alpha g_j(\theta^j_k)
\end{equation}
where $Nb(j)$ indicates the neighborhood of agent $j$, $\alpha$ is the step size, $g_j(\theta^j_k)$ is the (stochastic) gradient of $f_j$ at $\theta^j_k$, implemented by drawing a minibatch of sampled data points. More precisely, 
$g_j(\theta^j_k)=\frac{1}{b'}\sum_{q'\in \mathcal{D'}}\nabla f^{q'}_j(\theta^j_k),$
where $b'$ is the size of the minibatch $\mathcal{D'}$ selected uniformly at random from the data subset $\mathcal{D}_j$ available to Agent $j$.
\section{Proposed Algorithms}
\label{proposed_algorithm}

State-of-the-art algorithms such as CDSGD alternate between the \emph{gradient update} and \emph{consensus} steps. We propose two natural extensions where one can control the emphasis on \emph{consensus} relative to the \emph{gradient update} and hence, leads to interesting trade-offs between consensus and optimality.

\subsection{Increasing consensus}

Observe that the concatenated form of the CDSGD updates, ~\eqref{update_law}, can be expressed as
$$\Theta_{k+1} = (\Pi\otimes I_d)\Theta_k - \alpha\mathbf{g}(\Theta_k).$$
If we perform $\tau$ consensus steps interlaced with each gradient update, we can obtain the following concatenated form of the iterations of the parameter estimates:
\begin{equation}\label{con_update_icdsgd}
\Theta_{k+1} = (\Pi^\tau\otimes I_d)\Theta_k - \alpha\mathbf{g}(\Theta_k)
\end{equation}
where, 
$\mathbf{g}(\Theta_k) = \bigg[g_1^T(\theta^1_k),\;g_2^T(\theta^2_k), \dots, g_N^T(\theta^N_k)\bigg]^T.$
We call this variant \textit{incremental} consensus-based distributed SGD (i-CDSGD) which is detailed in Algorithm~\ref{ICDSGD}. Note, in a distributed setting, the this algorithm incurs an additional factor $\tau$ in communication complexity. 

A \emph{different} and more direct approach to control the trade-off between consensus and gradient would be as follows: 
\begin{equation}\label{gcd-sgd}
\Theta_{k+1} = (1-\omega)(\Pi\otimes I_d)\Theta_k + \omega(\Theta_k - \alpha\mathbf{g} .(\Theta_k))
\end{equation}
where, $0 \textless \omega \leq 1$ is a user-defined parameter. We call this algorithm \emph{generalized} consensus-based distributed SGD (g-CDSGD), and the full procedure is detailed in Algorithm~\ref{GCDSGD}.\\
\noindent
\begin{minipage}{.55\textwidth}
\vspace{0pt}
\begin{algorithm}[H]
    \caption{i-CDSGD}
    \begin{algorithmic}[1]\label{ICDSGD}
    \STATE \textbf{Initialization}: $\theta_0^j,v_0^j,\;j=1,2,...,N$, $\alpha$, $N$, $\tau$, $m$, $\Pi$
    \STATE Distribute the training data set to $N$ agents
    \FOR{\textit{each agent}}
      \STATE {Randomly shuffle each data subset}
      \FOR{$k=0:m$}
        \STATE {$t=0$}
        \FOR{$j = 1,...,N$}
          \STATE {$\theta^j_t = \theta^j_k$}
		\ENDFOR
        \FOR{$j=1,...,N$}
          \WHILE {$t\leq \tau-1$}
            \STATE {$\theta^j_{t+1} = \sum_{l\in Nb(j)}\pi_{jl}\theta^l_t$}           \tcp{\textbf{Incremental Consensus}}
			\STATE {$t = t+1$}
          \ENDWHILE
        \ENDFOR
        \STATE {$\hat{\theta} = \theta^j_t$}
        \STATE {$\theta^j_{k+1} = \hat{\theta} - \alpha g_j(\theta^j_k)$}
		\ENDFOR
    \ENDFOR
    \end{algorithmic}
\end{algorithm}
\end{minipage}
\begin{minipage}{.45\textwidth}
	\vspace{-110pt}
	\begin{algorithm}[H]
		\caption{g-CDSGD}
		\begin{algorithmic}[1]\label{GCDSGD}
			\STATE \textbf{Initialization}: $\omega, \theta_0^j,v_0^j,\;j=1,2,...,N$, $\alpha$, $N$, $m$, $\Pi$
			\STATE Distribute the training data set to $N$ agents
			\FOR{\textit{each agent}}
			\STATE {Randomly shuffle each data subset}
			\FOR{$k=0:m$}
			\STATE {$\hat{\theta}=\sum_{l\in Nb(j)}\pi_{jl}\theta^l_k$}
			\STATE {$\theta^j_{k+1} = (1-\omega)\hat{\theta}+ \omega(\theta^j_k - \alpha g_j(\theta^j_k))$}			\tcp{\textbf{Generalized Consensus}}
			\ENDFOR
			\ENDFOR
		\end{algorithmic}
	\end{algorithm}
\end{minipage}

By examining Eq.~\ref{gcd-sgd}, we observe that when $\omega$ approaches 0, the update law boils down to a only consensus protocol, and that when $\omega$ approaches 1, the method reduces to standard stochastic gradient descent (for individual agents). 

Next, we introduce the \emph{Nesterov momentum} variants of our aforementioned algorithms. The momentum term is typically used for speeding up the convergence rate with high momentum constant close to 1~\cite{sutskever2013importance}. 
For the purpose of reference and convenience, we embed the momentum variants of i-CDSGD and g-CDSGD within the Algorithms~\ref{ICDSGD} and~\ref{GCDSGD}. More details can be found in Algorithms~\ref{ICDMSGD} and~\ref{GCDMSGD} in the Supplementary Section~\ref{A1}.

\subsection{Tools for convergence analysis}

We now analyze the convergence of the iterates $\{\theta^j_k\}$ generated by our algorithms. 
Specifically, we identify an appropriate Lyapunov function (that is bounded from below) for each algorithm that decreases with each iteration, thereby establishing convergence. 
In our analysis, we use the concatenated (Kronecker) form of the updates.  For simplicity, let $\mathbf{P}=\Pi\otimes I_d\in\mathbb{R}^{Nd\times Nd}$. 

We begin the analysis for g-CDSGD by constructing a Lyapunov function that combines the true objective function with a regularization term involving a quadratic form of consensus as follows:
\begin{equation}\label{lyapunov}
V(\Theta):=\omega\mathcal{F}(\Theta) + \frac{1-\omega}{2\alpha}\Theta^T(I_{Nd} - \mathbf{P})\Theta
\end{equation}
\vspace{-2pt}
It is easy to show that $\sum_{j=1}^{N}f_j(\theta^j)$ is  $\gamma_m := max_j\{\gamma_j\}$-smooth, and that $V(\Theta)$ is $\hat{\gamma}$-smooth with 
\vspace{-2pt}
\begin{align*}
\hat{\gamma}&:=\omega\gamma_m+(1-\omega)\alpha^{-1}\lambda_{\text{max}}(I_{Nd}-\mathbf{P}) \\
&= \omega\gamma_m+(1-\omega)\alpha^{-1}(1-\lambda_N).
\vspace{-2pt}
\end{align*} 
Likewise, it is easy to show that $\sum_{j=1}^{N}f_j(\theta^j)$ is $H_m := min_j\{H_j\}$-strongly convex; therefore $V(\Theta)$ is $\hat{H}$-strongly convex with 
\begin{align*}
\hat{H} &:= \omega H_m+(1-\omega)(2\alpha)^{-1}\lambda_{\text{min}}(I_{Nd}-\mathbf{P}) \\
&= \omega H_m+(1-\omega)(2\alpha)^{-1}(1-\lambda_2).
\end{align*}
We also assume that there exists a lower bound $V_{\inf}$ for the function value sequence $\{V(\Theta_k)\}, \forall k$. When the objective functions are strongly convex, we have $V_{\inf} = V(\Theta^*)$, where $\Theta^*$ is the optimizer. 

Due to Assumptions~\ref{assump_1} and~\ref{assump_2}, it is straightforward to obtain an equivalence between the gradient of Eq.~\ref{lyapunov} and the update law of g-CDSGD.
Rewriting~\eqref{gcd-sgd}, we get:
\begin{equation}
\Theta_{k+1} = (1-\omega)\mathbf{P}\Theta_k + \omega(\Theta_k - \alpha\mathbf{g}(\Theta_k))
\end{equation}
Therefore, we obtain:
\begin{equation}
\begin{aligned}
\Theta_{k+1} &= \Theta_k - \Theta_k + (1-\omega)\mathbf{P} \Theta_k + \omega(\Theta_k - \alpha\mathbf{g}(\Theta_k))\\
             &= \Theta_k - \alpha\omega\Theta_k - (1-\omega)I_{Nd}\Theta_k + (1-\omega)\mathbf{P}\Theta_k\\
             &= \Theta_k - \alpha\underbrace{(\omega\mathbf{g}(\Theta_k) + \frac{1}{\alpha}(1-\omega)(I_{Nd} - \mathbf{P})\Theta_k)}_{\text{Lyapunov Gradient}}
\end{aligned}
\label{eq:lyap}
\end{equation}


The last term in \eqref{eq:lyap} is precisely the gradient of $V(\Theta)$. In the stochastic setting, $\mathbf{g}(\Theta_k)$ can be approximated by sampling one data point (or a mini-batch of data points) and the stochastic Lyapunov gradient is denoted by $\mathcal{S}(\Theta_k), \forall k$.

Similarly, the update laws for our proposed Nesterov momentum variants can be compactly analyzed using the above Lyapunov function. First, we rewrite the updates for g-CDMSGD as follows:
\begin{subequations}\label{gcdm}
\begin{alignat}{2}
  &\mathbf{y}_{k+1} = \Theta_k + \mu(\Theta_k - \Theta_{k-1})\\
  &\Theta_{k+1} = (1-\omega)\mathbf{P}\mathbf{y}_{k+1} + \omega(\mathbf{y}_{k+1} - \alpha\mathbf{g}(\mathbf{y}_{k+1}))
\end{alignat}
\end{subequations}
With a few algebraic manipulations, we get:
\begin{equation}
\begin{aligned}
&\Theta_{k+1} = \mathbf{y}_{k+1} - \mathbf{y}_{k+1} + (1-\omega)\mathbf{P}\mathbf{y}_{k+1}\\
&+\omega(\mathbf{y}_{k+1} -\alpha \mathbf{g}(\mathbf{y}_{k+1}))\\
&= \mathbf{y}_{k+1} - \alpha(\omega\mathbf{g}(\mathbf{y}_{k+1}) + \frac{1-\omega}{\alpha}(I_{Nd} - \mathbf{P})\mathbf{y}_{k+1})\\
\end{aligned}
\end{equation}
The above derivation simplifies the Nesterov momentum-based updates into a regular form which is more convenient for convergence analysis. For clarity, we separate this into two sub-equations. Let $\mathcal{S}(\mathbf{y}_{k+1}) = \omega\mathbf{g}(\mathbf{y}_{k+1})+\frac{1-\omega}{\alpha}(I_{Nd}-\mathbf{P})\mathbf{y}_{k+1}$. Thus, the updates for g-CDMSGD can be expressed as
\begin{subequations}\label{gcdm_1}
\begin{alignat}{2}
  &\mathbf{y}_{k+1} = \Theta_k + \mu(\Theta_k - \Theta_{k-1})\\
  &\Theta_{k+1} = \mathbf{y}_{k+1} - \alpha S(\mathbf{y}_{k+1}),
\end{alignat}
\end{subequations}
Please find the similar transformation for i-CDMSGD in Supplementary Section~\ref{A1}.

For analysis, we require a bound on the variance of the stochastic Lyapunov gradient $\mathcal{S}(\Theta_k)$ such that the variance of the gradient noise\footnote{As our proposed algorithm is a distributed variant of SGD, the noise in the performance is caused by the random sampling~\cite{song2015learning}.} can be bounded from above. The variance of $\mathcal{S}(\Theta_k)$ is defined as:
$$Var[\mathcal{S}(\Theta_k)]:=\E[\|\mathcal{S}(\Theta_k)\|^2]-\|\E[\mathcal{S}(\Theta_k)]\|^2.$$
The following assumption is standard in SGD convergence analysis, and is based on~\cite{bottou2016optimization}. 
\begin{assumption}\label{assump_3}
a) There exist scalars $r_2\geq r_1\textgreater 0$ such that $\nabla V(\Theta_k)^T\E[\mathcal{S}(\Theta_k)]\geq r_1\|\nabla V(\Theta_k)\|^2$ and $\|\E[ \mathcal{S}(\Theta_k)]\|\leq r_2\|\nabla V(\Theta_k)\|$ for all $k\in \N$; b) There exist scalars $B\geq 0$ and $B_V\geq 0$ such that $Var[\mathcal{S}(\Theta_k)]\leq B+B_V\|\nabla V(\Theta_k)\|^2$ for all $k\in\N$.
\end{assumption}

\begin{rem}
\label{rem1}
While Assumption~\ref{assump_3}(a) guarantees the sufficient descent of $V$ in the direction of $-\mathcal{S}(\Theta_k)$, Assumption~\ref{assump_3}(b) states that variance of $\mathcal{S}(\Theta_k)$ is bounded above by the second moment of $\nabla V(\Theta_k)$. The constant $B$ can be regarded to represent the second moment of noise involving in the gradient $ \mathcal{S}(\Theta_k)$. Therefore, the second moment of $ \mathcal{S}(\Theta_k)$ can be bounded above as $\E[\|\mathcal{S}(\Theta_k)\|^2]\leq B+B_m\|\nabla V(\Theta_k)\|^2$, where $B_m := B_V+r_2^2\geq r_1^2\textgreater 0$.
\end{rem}

For convergence analysis, we assume:
\begin{assumption}\label{assump_4}
There exists a constant $G\textgreater 0$ such that $\|\nabla V(x)\|\leq G, \forall x\in\mathbb{R}^d$.
\end{assumption}
As the Lyapunov function is a composite function with the true cost function which is Lipschitz continuous and the regularization term associated with consensus, it can be immediately obtained that $\|\nabla V(x)\|$ is bounded above by some positive constant.

Before turning to our main results, we present two auxiliary technical lemmas. 
\begin{lem}\label{lemma1}
Let Assumptions~\ref{assump_1} and~\ref{assump_2} hold. The iterates of g-CDSGD (Algorithm~\ref{GCDSGD}) satisfy the following inequality $\forall k\in \mathbb{N}$:
\begin{equation}
\begin{aligned}
\mathbb{E}&[V(\Theta_{k+1})]-V(\Theta_k) \\
&\leq -\alpha\nabla V(\Theta_k)^T\mathbb{E}[\mathcal{S}(\Theta_k)]+\frac{\hat{\gamma}}{2}\alpha^2\mathbb{E}[\|\mathcal{S}(\Theta_k)\|^2] .
\end{aligned}
\end{equation}
\end{lem}
\begin{lem}\label{lemma2}
Let Assumptions~\ref{assump_1},~\ref{assump_2}, and~\ref{assump_3} hold. The iterates of g-CDSGD (Algorithm~\ref{GCDSGD}) satisfy the following inequality $\forall k\in \mathbb{N}$:
\begin{equation}\label{lem2}
\begin{aligned}
\E&[V(\Theta_{k+1})]-V(\Theta_k)\\
&\leq -(r_1-\frac{\hat{\gamma}}{2}\alpha B_m)\alpha\|\nabla V(\Theta_k)\|^2
                       +\frac{\hat{\gamma}}{2}\alpha^2B .
\end{aligned}
\end{equation}
\end{lem}
We provide the proof of Lemmas~\ref{lemma1} and~\ref{lemma2} in the Supplementary Section~\ref{supp}.
To guarantee that the first term on the right hand side is strictly negative, the step size $\alpha$ should be chosen such that
\begin{equation}\label{step_size}
0\textless \alpha \leq \frac{r_1-(1-\omega)(1-\lambda_N^\tau)B_m}{\omega B_m\gamma_m}.
\end{equation}
\section{Analysis and Main Results}
\label{main_results}

This section presents the main results by analyzing the convergence properties of the proposed algorithms. Our main results are grouped as follows: 
(i) we provide rigorous convergence analysis for g-CDSGD and i-CDSGD for both strongly convex and non-convex objective functions. 
(ii) we analyze their momentum variants only for strongly convex objective functions. It is noted that all proofs are provided in the Supplementary Section~\ref{A1}.
\subsection{Convergence Analysis for i-CDSGD and g-CDSGD}

Our analysis will consist of two components: establishing an upper bound on how far away the estimates of the individual agents are with respect to their empirical mean  (which we call the \emph{consensus bound}), and establishing an upper bound on how far away the overall procedure is with respect to the optimum (which we call the \emph{optimality bound}).

First, we obtain consensus bounds for the g-CDSGD and i-CDSGD as follows.
\begin{proposition}{(Consensus with fixed step size, g-CDSGD)}\label{prop1}
Let Assumptions~\ref{assump_1},~\ref{assump_2},~\ref{assump_4} hold. The iterates of g-CDSGD (Algorithm~\ref{GCDSGD}) satisfy the following inequality $\forall k\in \mathbb{N}$, when $\alpha$ satisfies Eq.~\ref{step_size},
\begin{equation}
\E[\|\theta^j_k - s_k\|]\leq \frac{\omega\alpha h}{1-\hat{\lambda}_2}
\end{equation}
where $s_k = \frac{1}{N}\sum_{j=1}^{N}\theta_k^j$, $\hat{\lambda}_2$ is the second-largest eigenvalue of the matrix $\mathbf{Q}=(1-\omega)\mathbf{P}+\omega I_{Nd}$.
\end{proposition}
\begin{proposition}{(Consensus with fixed step size, i-CDSGD)}\label{prop2}
Let Assumptions~\ref{assump_1},~\ref{assump_2},~\ref{assump_4} hold. The iterates of i-CDSGD (Algorithm~\ref{ICDSGD}) satisfy the following inequality $\forall k\in \mathbb{N}$, when $\alpha$ satisfies $0\textless\alpha\leq\frac{r_1-(1-\lambda_N^\tau)B_m}{\gamma_mB_m}$,
\begin{equation}
\E[\|\theta^j_k - s_k\|]\leq \frac{\alpha h}{1-\lambda_2^\tau}
\end{equation}
\end{proposition}

We provide a discussion on comparing the consensus bounds in the Supplementary Section~\ref{A2}. 
Next, we obtain optimality bounds for g-CDSGD and i-CDSGD. 
\begin{thm}\label{thm1}{(Convergence of g-CDSGD in strongly convex case)}
Let Assumptions~\ref{assump_1},~\ref{assump_2} and~\ref{assump_3} hold. When the step size satisfies Eq.~\ref{step_size}, the iterates of g-CDSGD (Algorithm~\ref{GCDSGD}) satisfy the following inequality $\forall k\in \mathbb{N}$:
\begin{equation}
\E[D_k]\leq C_1^{k-1}D_1+C_2\sum_{q=0}^{k-1}C_1^{q}
\end{equation}
where $D_k =V(\Theta_k)-V^*, C_1 = 1-(\omega\alpha H_m+\frac{1-\omega}{2}(1-\lambda_2))r_1, C_2 = \frac{(\alpha^2\gamma_m\omega+\alpha(1-\omega)(1-\lambda_N))B}{2}$.
\end{thm}
\begin{thm}\label{thm2}{(Convergence of i-CDSGD in strongly convex case)}
Let Assumptions~\ref{assump_1},~\ref{assump_2} and~\ref{assump_3} hold. When the step size satisfies $0\textless\alpha\leq\frac{r_1-(1-\lambda_N^\tau)B_m}{\gamma_mB_m}$, the iterates of i-CDSGD (Algorithm~\ref{ICDSGD}) satisfy the following inequality $\forall k\in \mathbb{N}$:
\begin{equation}
\E[D_k]\leq C_3^{k-1}D_1+C_4\sum_{q=0}^{k-1}C_3^{q}
\end{equation}
where $D_k =V(\Theta_k)-V^*$, $C_3 = 1-(\alpha H_m+\frac{1}{2}(1-\lambda_2^\tau))r_1$, $C_4 = \frac{(\alpha^2\gamma_m+\alpha(1-\lambda_N^\tau))B}{2}$.
\end{thm}
Although we show the convergence for strongly convex objectives, we note that objective functions are higly non-convex for most deep learning applications. While convergence to a global minimum in such cases is extremely difficult to establish, we prove that g-CDSGD and i-CDSGD still exhibits weaker (but meaningful) notions of convergence.

\begin{thm}\label{thm3}{(Convergence to the first-order stationary point for non-convex case of g-CDSGD)}
Let Assumptions~\ref{assump_1},~\ref{assump_2}, and~\ref{assump_3} hold. When the step size satisfies Eq.~\ref{step_size}, the iterates of g-CDSGD (Algorithm~\ref{GCDSGD}) satisfy the following inequality $\forall K\in \mathbb{N}$:
\begin{equation}\label{part1_thm2}
\begin{aligned}
\E[&\frac{1}{K}\sum_{k=1}^{K}\|\nabla V(\Theta_k)\|^2] \\
&\leq\frac{(\omega\gamma_m\alpha+(1-\omega)(1-\lambda_N))B}{r_1}+\frac{2(V(\Theta_1)-V_{\text{inf}})}{Kr_1\alpha}
\end{aligned}
\end{equation}
\end{thm}

\begin{thm}\label{thm4}{(Convergence to the first-order stationary point for non-convex case of i-CDSGD)}
Let Assumptions~\ref{assump_1},~\ref{assump_2}, and~\ref{assump_3} hold. When the step size satisfies $0\textless\alpha\leq\frac{r_1-(1-\lambda_N^\tau)B_m}{\gamma_mB_m}$, the iterates of i-CDSGD (Algorithm~\ref{ICDSGD}) satisfy the following inequality $\forall K\in \mathbb{N}$:
\begin{equation}\label{part1_thm2}
\begin{aligned}
\E[&\frac{1}{K}\sum_{k=1}^{K}\|\nabla V(\Theta_k)\|^2] \\
&\leq\frac{(\gamma_m\alpha+(1-\lambda_N^\tau))B}{r_1}+\frac{2(V(\Theta_1)-V_{\text{inf}})}{Kr_1\alpha} .
\end{aligned}
\end{equation}
\end{thm}

\begin{rem}
Let us discuss the rates of convergence suggested by Theorems~\ref{thm1} and~\ref{thm3}. We observe that when the objective function is strongly convex, the function value sequence $\{V(\Theta_k)\}$ can \textit{linearly} converge to within a fixed radius of convergence, which can be calculated as follows: 
\[
\lim_{k\to\infty}\mathbb{E}[V(\Theta_k)-V*]\leq\frac{B[\omega\alpha\gamma_m+(1-\omega)(1-\lambda_N)]}{2r_1(\omega H_m+\alpha^{-1}(1-\omega)(1-\lambda_2))}.
\] 
When the objective function is non-convex, we cannot claim linear convergence. However, Theorem~\ref{thm3} asserts that the average of the second moment of the Lyapunov gradient is bounded from above. Recall that the parameter $B$ bounds the variance of the ``noise'' due to the stochasticity of the gradient, and if $B = 0$, Theorem~\ref{thm3} implies that $\{\Theta_k\}$ asymptotically converges to a first-order stationary point.
\end{rem}

\begin{rem}
For g-CDSGD, let us investigate the corner cases where $\omega\to 0$ or $\omega\to 1$.  For the strongly convex case, when $\omega\to 1$, we have $\frac{\alpha cB}{2r_1}$, where $c = \frac{\gamma_m}{H_m}$ is the condition number. This suggests that if consensus is not a concern, then
each iterate $\{\theta^j_k\}$ converges to its own respective $\theta^j_*$, as depicted in Fig.~\ref{Figure1}. On the other hand, when $\omega\to 0$, the upper bound converges to $\frac{\alpha B(1-\lambda_N)}{2r_1(1-\lambda_2)}$. In such a scenario, 
each agent sufficiently communicates its own information with other agents to arrive at an agreement. In this case, the upper bound depends on the topology of the communication network. If $\lambda_N\approx 0$, this results in:
\[
\lim_{k\to \infty}\mathbb{E}[V(\Theta_k)-V^*]\leq \frac{B\alpha}{2r_1(1-\lambda_2)}.
\] 

For the non-convex case, when $\omega\to 1$, the upper bound suggested by Theorem~\ref{thm3} is $\frac{\alpha\gamma_mB}{r_1}$, while $\omega\to 0$ leads to $\frac{(1-\lambda_N)B}{r_1}$, which is roughly $\frac{B}{r_1}$ if $\lambda_N\approx 0$. 
\end{rem}

We also compare i-CDSGD and CDSGD with g-CDSGD in terms of the optimality upper bounds to arrive at a suitable lower bound for $\omega$. However, due to the space limit, the analysis is presented in the Supplementary Section~\ref{A2}. 

\subsection{Convergence Analysis for momentum variants}
We next provide a convergence analysis for the g-CDMSGD algorithm, summarized in the update laws given in Eq.~\ref{gcdm_1}. A similar analysis can be applied to i-CDMSGD. Before stating the main result, we define the sequence $\phi_k(\Theta),~k = 1,2,\ldots$ as:
\[
\phi_1(\Theta) = V(\Theta_1) + \frac{\hat{H}}{2}\|\Theta - \Theta_1\|^2 ,~\text{and}
\]
\begin{equation}\label{estimate}
\begin{aligned}
\phi_{k+1} =&(1-\sqrt{\hat{H}\alpha})\phi_k(\Theta) + \sqrt{\hat{H}\alpha}(\hat{V}(\mathbf{y}_k) \\
&+(\mathcal{S}_k,\Theta - \mathbf{y}_k) +\frac{\hat{H}}{2}\|\Theta - \mathbf{y}_k\|^2
\end{aligned}
\end{equation}
where $\hat{V}$ represents the average of the objective function values of a mini-batch.
We define $\phi^*_k$ as follows
\[
\phi^*_k = \min\limits_{\Theta\in\mathbb{R}^{Nd}}\phi_k(\Theta)
\]

Further, from Assumption~\ref{assump_3}, we see that $Var[\mathcal{S}(\mathbf{y}_k)]\leq B+B_{V}\|\nabla V(\mathbf{y}_k)\|^2$. Combining Assumption~\ref{assump_4} and $\text{Var}[\mathcal{S}(\mathbf{y}_k)] :=\mathbb{E}[\|\mathcal{S}(\mathbf{y}_k) - \nabla V(\mathbf{y}_k)\|^2]$, we have $\mathbb{E}[\|\mathcal{S}(\mathbf{y}_k) - \nabla V(\mathbf{y}_k)\|^2]\leq B+B_{V}G^2$. 

We now state our main result, which characterizes the performance of g-CDMSGD. To our knowledge, this is the first theoretical result for momentum-based versions of consensus-distributed SGD. 
\begin{thm}\label{thm5}{(Convergence of g-CDMSGD, strongly convex case)}
Let Assumptions~\ref{assump_1},~\ref{assump_2},~\ref{assump_3}, and~\ref{assump_4} hold. If the step size satisfies $\alpha\leq\textnormal{min}\{\frac{r_1-(1-\omega)(1-\lambda_N)B_m}{\omega B_m\gamma_m}, \frac{1}{\hat{H}}, \frac{1}{2\hat{\gamma}}\}$, we have:
\begin{equation}
\begin{aligned}
\mathbb{E}[V(\Theta_k) - V^*]\leq (1-&\sqrt{\hat{H}\alpha})^{k-1}(\phi_1^*-V^*)\\
&+\sqrt{\frac{\alpha}{\hat{H}}}(B+B_{V}G^2) .
\end{aligned}
\end{equation}
\end{thm}
\begin{thm}\label{thm6}{(Convergence of i-CDMSGD, strongly convex case)}
Let Assumptions~\ref{assump_1},~\ref{assump_2},~\ref{assump_3}, and~\ref{assump_4} hold. If the step size satisfies $\alpha\leq\textnormal{min}\{\frac{r_1-(1-\lambda_N^\tau)B_m}{B_m\gamma_m}, \frac{1}{\hat{H}}, \frac{1}{2\hat{\gamma}}\}$, we have:
\begin{equation}
\begin{aligned}
\mathbb{E}[V(\Theta_k) - V^*]\leq (1-&\sqrt{\hat{H}\alpha})^{k-1}(\phi_1^*-V^*)\\
&+\sqrt{\frac{\alpha}{\hat{H}}}(B+B_{V}G^2) .
\end{aligned}
\end{equation}
\end{thm}
Note, although the theorem statements look the same for g-CDMSGD and i-CDMSGD, the constants $\hat{H}$ are significantly different from each other. Theorem~\ref{thm5} suggests that with a sufficiently small step size, using Nesterov acceleration results in a linear convergence (with parameter $1-\sqrt{\hat{H}\alpha}$) up to a neighbourhood of $V^*$ of radius $\sqrt{\frac{\alpha}{\hat{H}}}(B+B_VG^2)$. When $k\to \infty$, the first term on the right hand side vanishes and substituting $\hat{H}=\omega H_m+(1-\omega)(2\alpha)^{-1}(1-\lambda_2)$ into $\sqrt{\frac{\alpha}{\hat{H}}}(B+B_VG^2)$, we have
\[
\sqrt{\frac{\alpha}{\omega H_m+(1-\omega)(2\alpha)^{-1}(1-\lambda_2)}}(B+B_VG^2),
\]
which implies that the upper bound is related to the spectral gap $1-\lambda_2$ of the network; hence, a similar conclusion Theorem~\ref{thm1} can be deduced. When $\omega \to 0$, the upper bound becomes $\alpha\sqrt{\frac{1}{2(1-\lambda)}}(B+B_VG^2)$. However, $\omega \to 1$ leads to $\sqrt{\frac{\alpha}{H_m}}(B+B_VG^2)$. These two scenarios demonstrates that the ``gradient noise'' cased by the stochastic sampling negatively affects the convergence. One can use $\omega$ to trade-off the consensus and updates. 

Next, we discuss the upper bounds obtained when $k\to \infty $ for g-CDSGD and g-CDMSGD. (1) $\omega\to 0$: When $B_V$ is sufficiently small and $r_1\approx \frac{1}{2\sqrt{2}}$, it can be observed that the optimality bound for the Nesterov momentum variant is smaller than that for g-CDSGD as $\frac{B\alpha}{1-\lambda_2}\textgreater \frac{B\alpha}{\sqrt{1-\lambda_2}}$; (2) $\omega\to 1$: When $\gamma_m$ and $r_1$ are carefully selected such that $\frac{\gamma_m}{2r_1}\approx1$, we have $B\sqrt{\frac{\alpha}{H_m}}\textless\frac{B\alpha}{H_m}$ when $\frac{\alpha}{H_m}>1$. Therefore, introducing the momentum can speed up the convergence rate with appropriately chosen hyperparameters.

\begin{figure*}[!t]
	\centering
\begin{subfigure}{.47\textwidth}
	\centering
	\includegraphics[width=0.95\linewidth,trim={0.5cm, 0.2cm, 0.1cm, 1.1cm},clip]{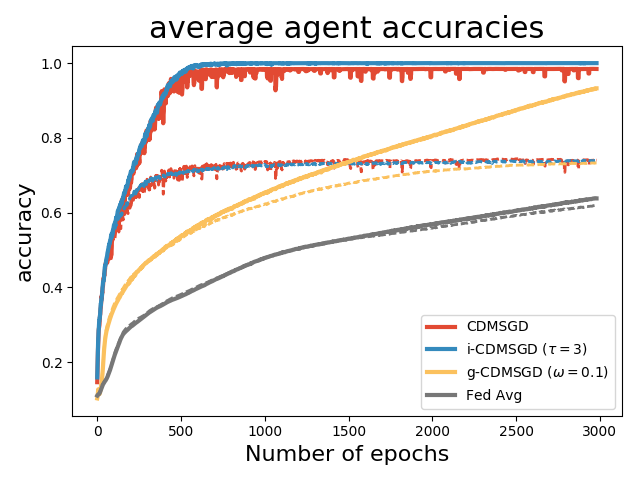}
	\caption{Performance of different algorithms with unbalanced sample distribution among agents. (Dashed lines represent test accuracy \&  solid lines represent training accuracy.)} 
	\label{Fig:2a}
\end{subfigure}
\hfill
\begin{subfigure}{.47\textwidth}
	\centering
	\includegraphics[width=0.95\linewidth,trim={0.5cm, 0.3cm, 0.0cm, 1.0cm},clip]{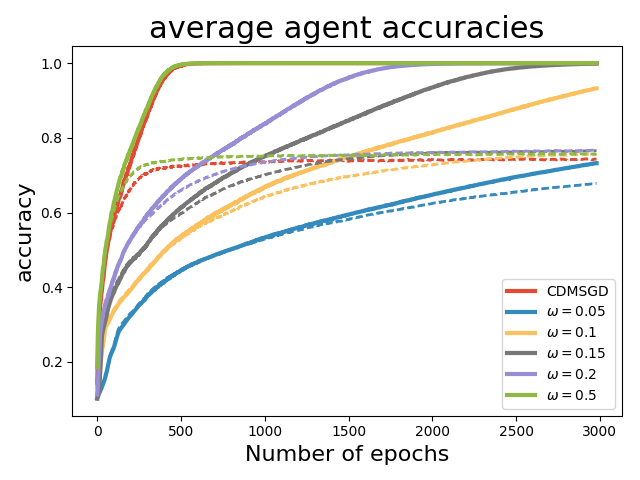}
	\caption{Performance of g-CDMSGD for different $\omega$ values. (Dashed lines represent test accuracy \& solid lines represent training accuracy.) }
	\label{Fig:2b} 
\end{subfigure}

\begin{subfigure}{.47\textwidth}
	\centering
	\includegraphics[width=0.95\linewidth,trim={0.6cm, 0.5cm, 1cm, 0.35cm},clip]{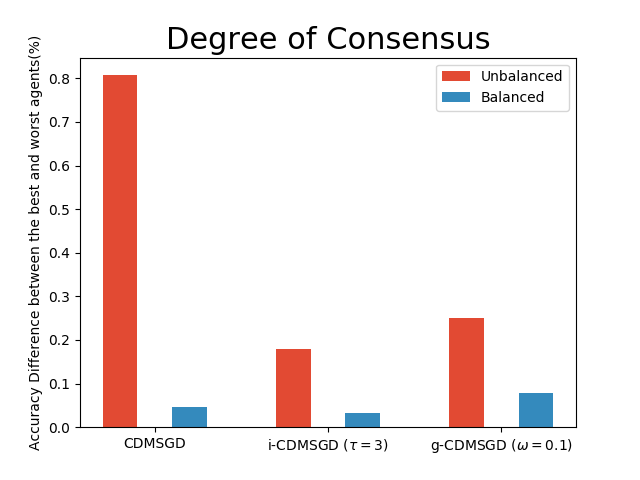}\vspace{-5pt}
	\caption{The accuracy percentage difference between the best and the worst agents for different algorithms with unbalanced and balanced sample distribution among agents.}
	\label{Fig:2c}
\end{subfigure}
\hfill
\begin{subfigure}{.47\textwidth}
	\centering
	\includegraphics[width=0.95\linewidth,trim={0.5cm, 0.5cm, 1cm, 0.7cm},clip]{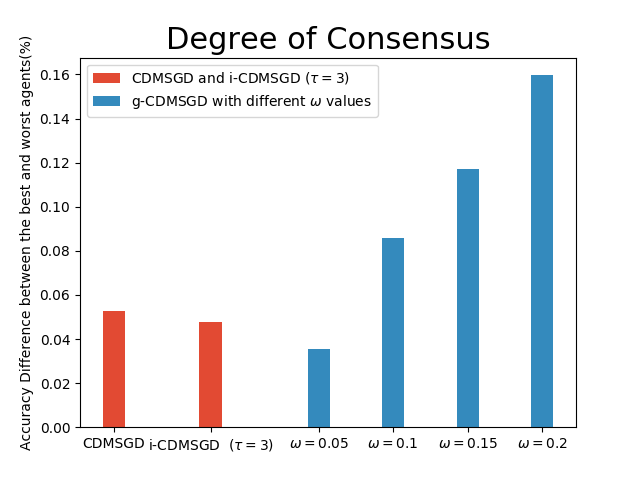}\vspace{-5pt}
	\caption{The accuracy percentage difference between the best and the worst agents with balanced sample distribution for CDMSGD, i-CDMSGD and g-CDMSGD (varying $\omega$.)}\vspace{-0pt}
	\label{Fig:2d}
\end{subfigure}
\caption{Results from numerical experiments with CIFAR-10 data set}
\label{Fig:results}
\end{figure*}





\section{Experimental Results}\label{experimental_results}

We validate our algorithms via several experimental results using the CIFAR-10 image recognition dataset (with standard training and testing sets). The model adopted for the experiments is a deep convolutional neural network (CNN) (with ReLU activations) which includes 2 convolutional layers with 32 filters each followed by a max pooling layer, then 2 more convolutional layers with 64 filters each followed by another max pooling layer, and a dense layer with 512 units. The mini-batch size is set to 512, and step size is set to 0.01 in all experiments. All experiments were performed using Keras with TensorFlow~\cite{chollet2015keras,abadi2016tensorflow}. 
We use a sparse network topology with 5 agents. 
We use both balanced and unbalanced data sets for our experiments. In the balanced case, agents have an equal share of the entire training set. However, in the unbalanced case, agents have (randomly selected) unequal parts of the training set while making sure that each agent has at least half of the equal share amount of examples. We summarize our key experimental results in this section, with more details and results provided in the Supplementary Section~\ref{A4}.

\textbf{Performance of algorithms}. 
In Figure~\ref{Fig:2a}, we compare the performance of the momentum variants of our proposed algorithms, i-CDMSGD and g-CDMSGD (with $\omega = 0.1$) with state-of-the art techniques such as CDMSGD and Federated Averaging using an unbalanced data set. All algorithms were run for 3000 epochs. Observing the average accuracy over all the agents for both training and test data, we note that i-CDMSGD can converge as fast as CDMSGD with lesser fluctuation in the performance across epochs. While being slower in convergence, g-CDMSGD acheves similar performance (with test data) with less fluctuation as well as smaller generalization gap (i.e., difference between training and testing accuracy). All algorithms significantly outperform Federated Averaging in terms of average accuracy. We also vary the tuning parameter $\omega$ for g-CDMSGD to show (in Figure~\ref{Fig:2b}) that it is able to achieve similar (or better) convergence rate as CDMSGD using higher $\omega$ values with some sacrifice in terms of the generalization gap. 

\textbf{Degree of Consensus}. One of the main contribution of our paper is to show that one can control the degree of consensus while maintaining average accuracy in distributed deep learning. We demonstrate this by observing the accuracy difference between the best and the worst performing agents (identified by computing the mean accuracy for the last 100 epochs). As shown in Figure~\ref{Fig:2c}, the degree of consensus is similar for all three algorithms for balanced data set, with i-CDMSGD performing slightly better than the rest. However, for an unbalanced set, both i-CDMSGD and g-CDMSGD perform significantly better compared to CDMSGD. Note, the degree of consensus can be further improved for g-CDMSGD using lower values of $\omega$ as shown in Figure~\ref{Fig:2d}. However, the convergence becomes relatively slower as shown in Figure~\ref{Fig:2b}. We do not compare these results with the Federated Averaging algorithm as it performs a brute force consensus at every epoch using centralized parameter server. We also do not vary $\tau$ as the doubly stochastic agent interaction matrix for the small agent population becomes stationary very quickly with a very small value of $\tau$. However, this will be explored in our future work with significantly bigger networks.
\vspace{-10pt}



\section{Conclusion and Future Work}\label{conclusion}\vspace{-5pt}
For investigating the trade-off between consensus and optimality in distributed deep learning with constrained communication topology, this paper presents two new algorithms, called i-CDSGD and g-CDSGD and their momentum variants. 
We show the convergence properties for the proposed algorithms and the relationships between the hyperparameters and the consensus \& optimality bounds. Theoretical and experimental comparison with the state-of-the-art algorithm called CDSGD, shows that i-CDSGD, and g-CDSGD can improve the degree of consensus among the agents while maintaining the average accuracy especially when there is data imbalance among the agents. 
Future research directions include learning with non-uniform data distributions among agents and time-varying networks. 

\clearpage
\bibliography{NonIID}

\begin{thebibliography}{10}

\bibitem{lecun2015deep}
Yann LeCun, Yoshua Bengio, and Geoffrey Hinton.
\newblock Deep learning.
\newblock {\em Nature}, 521(7553):436--444, 2015.

\bibitem{recht2011hogwild}
Benjamin Recht, Christopher Re, Stephen Wright, and Feng Niu.
\newblock Hogwild: A lock-free approach to parallelizing stochastic gradient
  descent.
\newblock In {\em Advances in Neural Information Processing Systems}, pages
  693--701, 2011.

\bibitem{jin2016scale}
Peter~H Jin, Qiaochu Yuan, Forrest Iandola, and Kurt Keutzer.
\newblock How to scale distributed deep learning?
\newblock {\em arXiv preprint arXiv:1611.04581}, 2016.

\bibitem{lenz2015deep}
Ian Lenz, Honglak Lee, and Ashutosh Saxena.
\newblock Deep learning for detecting robotic grasps.
\newblock {\em The International Journal of Robotics Research},
  34(4-5):705--724, 2015.

\bibitem{gubbi2013internet}
Jayavardhana Gubbi, Rajkumar Buyya, Slaven Marusic, and Marimuthu Palaniswami.
\newblock Internet of things (iot): A vision, architectural elements, and
  future directions.
\newblock {\em Future generation computer systems}, 29(7):1645--1660, 2013.

\bibitem{lane2015early}
Nicholas~D Lane, Sourav Bhattacharya, Petko Georgiev, Claudio Forlivesi, and
  Fahim Kawsar.
\newblock An early resource characterization of deep learning on wearables,
  smartphones and internet-of-things devices.
\newblock In {\em Proceedings of the 2015 International Workshop on Internet of
  Things towards Applications}, pages 7--12. ACM, 2015.

\bibitem{lane2015can}
Nicholas~D Lane and Petko Georgiev.
\newblock Can deep learning revolutionize mobile sensing?
\newblock In {\em Proceedings of the 16th International Workshop on Mobile
  Computing Systems and Applications}, pages 117--122. ACM, 2015.

\bibitem{dean2012large}
Jeffrey Dean, Greg Corrado, Rajat Monga, Kai Chen, Matthieu Devin, Mark Mao,
  Andrew Senior, Paul Tucker, Ke~Yang, Quoc~V Le, et~al.
\newblock Large scale distributed deep networks.
\newblock In {\em Advances in neural information processing systems}, pages
  1223--1231, 2012.

\bibitem{jiang2017collaborative}
Zhanhong Jiang, Aditya Balu, Chinmay Hegde, and Soumik Sarkar.
\newblock Collaborative deep learning in fixed topology networks.
\newblock {\em Neural Information Processing Systems (NIPS)}, 2017.

\bibitem{li2014communication}
Mu~Li, David~G Andersen, Alexander~J Smola, and Kai Yu.
\newblock Communication efficient distributed machine learning with the
  parameter server.
\newblock In {\em Advances in Neural Information Processing Systems}, pages
  19--27, 2014.

\bibitem{das2016distributed}
Dipankar Das, Sasikanth Avancha, Dheevatsa Mudigere, Karthikeyan Vaidynathan,
  Srinivas Sridharan, Dhiraj Kalamkar, Bharat Kaul, and Pradeep Dubey.
\newblock Distributed deep learning using synchronous stochastic gradient
  descent.
\newblock {\em arXiv preprint arXiv:1602.06709}, 2016.

\bibitem{lian2017can}
Xiangru Lian, Ce~Zhang, Huan Zhang, Cho-Jui Hsieh, Wei Zhang, and Ji~Liu.
\newblock Can decentralized algorithms outperform centralized algorithms? a
  case study for decentralized parallel stochastic gradient descent.
\newblock In {\em Advances in Neural Information Processing Systems}, pages
  5336--5346, 2017.

\bibitem{berahas2017balancing}
Albert~S Berahas, Raghu Bollapragada, Nitish~Shirish Keskar, and Ermin Wei.
\newblock Balancing communication and computation in distributed optimization.
\newblock {\em arXiv preprint arXiv:1709.02999}, 2017.

\bibitem{jiang2017generalised}
Zhanhong Jiang, Kushal Mukherjee, and Soumik Sarkar.
\newblock Generalised gossip-based subgradient method for distributed
  optimisation.
\newblock {\em International Journal of Control}, pages 1--17, 2017.

\bibitem{mcmahan2016communication}
H~Brendan McMahan, Eider Moore, Daniel Ramage, Seth Hampson, et~al.
\newblock Communication-efficient learning of deep networks from decentralized
  data.
\newblock {\em arXiv preprint arXiv:1602.05629}, 2016.

\bibitem{scaman2017optimal}
Kevin Scaman, Francis Bach, S{\'e}bastien Bubeck, Yin~Tat Lee, and Laurent
  Massouli{\'e}.
\newblock Optimal algorithms for smooth and strongly convex distributed
  optimization in networks.
\newblock In {\em International Conference on Machine Learning}, pages
  3027--3036, 2017.

\bibitem{qu2017accelerated}
Guannan Qu and Na~Li.
\newblock Accelerated distributed nesterov gradient descent.
\newblock {\em arXiv preprint arXiv:1705.07176}, 2017.

\bibitem{zhang2015deep}
Sixin Zhang, Anna~E Choromanska, and Yann LeCun.
\newblock Deep learning with elastic averaging sgd.
\newblock In {\em Advances in Neural Information Processing Systems}, pages
  685--693, 2015.

\bibitem{blot2016gossip}
Michael Blot, David Picard, Matthieu Cord, and Nicolas Thome.
\newblock Gossip training for deep learning.
\newblock {\em arXiv preprint arXiv:1611.09726}, 2016.

\bibitem{xu2017adaptive}
Zheng Xu, Gavin Taylor, Hao Li, Mario Figueiredo, Xiaoming Yuan, and Tom
  Goldstein.
\newblock Adaptive consensus admm for distributed optimization.
\newblock In {\em International Conference on Machine Learning}, pages
  3841--3850, 2017.

\bibitem{zheng2016asynchronous}
Shuxin Zheng, Qi~Meng, Taifeng Wang, Wei Chen, Nenghai Yu, Zhi-Ming Ma, and
  Tie-Yan Liu.
\newblock Asynchronous stochastic gradient descent with delay compensation for
  distributed deep learning.
\newblock In {\em International Conference on Machine Learning}, pages
  4120--4129, 2017.

\bibitem{zhang2017projection}
Wenpeng Zhang, Peilin Zhao, Wenwu Zhu, Steven~CH Hoi, and Tong Zhang.
\newblock Projection-free distributed online learning in networks.
\newblock In {\em International Conference on Machine Learning}, pages
  4054--4062, 2017.

\bibitem{wangni2017gradient}
Jianqiao Wangni, Jialei Wang, Ji~Liu, and Tong Zhang.
\newblock Gradient sparsification for communication-efficient distributed
  optimization.
\newblock {\em arXiv preprint arXiv:1710.09854}, 2017.

\bibitem{wen2017terngrad}
Wei Wen, Cong Xu, Feng Yan, Chunpeng Wu, Yandan Wang, Yiran Chen, and Hai Li.
\newblock Terngrad: Ternary gradients to reduce communication in distributed
  deep learning.
\newblock In {\em Advances in Neural Information Processing Systems}, pages
  1508--1518, 2017.

\bibitem{tsianos2012distributed}
Konstantinos~I Tsianos and Michael~G Rabbat.
\newblock Distributed strongly convex optimization.
\newblock In {\em Communication, Control, and Computing (Allerton), 2012 50th
  Annual Allerton Conference on}, pages 593--600. IEEE, 2012.

\bibitem{sutskever2013importance}
Ilya Sutskever, James Martens, George Dahl, and Geoffrey Hinton.
\newblock On the importance of initialization and momentum in deep learning.
\newblock In {\em International conference on machine learning}, pages
  1139--1147, 2013.

\bibitem{song2015learning}
Shuang Song, Kamalika Chaudhuri, and Anand Sarwate.
\newblock Learning from data with heterogeneous noise using sgd.
\newblock pages 894--902, 2015.

\bibitem{bottou2016optimization}
L{\'e}on Bottou, Frank~E Curtis, and Jorge Nocedal.
\newblock Optimization methods for large-scale machine learning.
\newblock {\em arXiv preprint arXiv:1606.04838}, 2016.

\bibitem{chollet2015keras}
Fran{\c{c}}ois Chollet et~al.
\newblock Keras, 2015.

\bibitem{abadi2016tensorflow}
Mart{\'\i}n Abadi, Paul Barham, Jianmin Chen, Zhifeng Chen, Andy Davis, Jeffrey
  Dean, Matthieu Devin, Sanjay Ghemawat, Geoffrey Irving, Michael Isard, et~al.
\newblock Tensorflow: A system for large-scale machine learning.
\newblock In {\em OSDI}, volume~16, pages 265--283, 2016.

\bibitem{nesterov2013introductory}
Yurii Nesterov.
\newblock {\em Introductory lectures on convex optimization: A basic course},
  volume~87.
\newblock Springer Science \& Business Media, 2013.

\bibitem{nitanda2014stochastic}
Atsushi Nitanda.
\newblock Stochastic proximal gradient descent with acceleration techniques.
\newblock In {\em Advances in Neural Information Processing Systems}, pages
  1574--1582, 2014.

\end{thebibliography}
\bibliographystyle{unsrt}

\clearpage
\appendix
\section{Supplementary Materials}\label{supp}

\subsection{Omitted algorithms}\label{A1}

\textbf{Update rules for the momentum variant, i-CDMSGD}. The compact form of i-CDMSGD is expressed as follows:
\begin{subequations}
\begin{alignat}{2}
  &\mathbf{y}_{k+1} = \Theta_k + \mu(\Theta_k - \Theta_{k-1})\\
  &\Theta_{k+1} = \mathbf{P}^\tau\mathbf{y}_{k+1}  - \alpha\mathbf{g}(\mathbf{y}_{k+1}) .
\end{alignat}
\end{subequations}
Rewriting the above equations yields:
\begin{equation}
\begin{aligned}
&\Theta_{k+1} = \mathbf{y}_{k+1} - \mathbf{y}_{k+1} + \mathbf{P}^\tau\mathbf{y}_{k+1} -\alpha \mathbf{g}(\mathbf{y}_{k+1})\\
&= \mathbf{y}_{k+1} - \alpha(\mathbf{g}(\mathbf{y}_{k+1}) + \frac{1}{\alpha}(I_{Nd} - \mathbf{P}^\tau)\mathbf{y}_{k+1}) .
\end{aligned}
\end{equation}
Letting $\mathcal{S}(\mathbf{y}_{k+1}) = \mathbf{g}(\mathbf{y}_{k+1}) + \frac{1}{\alpha}(I_{Nd} - \mathbf{P}^\tau)$, we have
\begin{subequations}
\begin{alignat}{2}
  &\mathbf{y}_{k+1} = \Theta_k + \mu(\Theta_k - \Theta_{k-1}) ,\\ 
  &\Theta_{k+1} = \mathbf{y}_{k+1}  - \alpha\mathcal{S}(\mathbf{y}_{k+1}) .
\end{alignat}
\end{subequations}

\subsubsection{Proofs of main lemmas}

We repeat the statements of all lemmas and theorems for completeness.

\noindent\textbf{Lemma~\ref{lemma1}}:
Let Assumptions~\ref{assump_1} and~\ref{assump_2} hold. The iterates of g-CDSGD (Algorithm~\ref{GCDSGD}) satisfy the following $\forall k\in \mathbb{N}$:
\begin{equation}
\begin{aligned}
\mathbb{E}[V(\Theta_{k+1})] &-V(\Theta_k)  \\
&\leq -\alpha\nabla V(\Theta_k)^T\mathbb{E}[\mathcal{S}(\Theta_k)]\\
                                       +\frac{\hat{\gamma}}{2}\alpha^2\mathbb{E}[\|\mathcal{S}(\Theta_k)\|^2] .
\end{aligned}
\end{equation}
\begin{proof}
By Assumption~\ref{assump_1}, the iterates generated by g-CDSGD satisfy:
\begin{equation}
\begin{aligned}
V(&\Theta_{k+1})-V(\Theta_k) \\
&\leq \nabla V(\Theta_k)^T(\Theta_{k+1}-\Theta_k)
+\frac{1}{2}\hat{\gamma} \|\Theta_{k+1}-\Theta_k\|^2 \\
&=-\alpha\nabla V(\Theta_k)^T\nabla \mathcal{S}(\Theta_k)+\frac{1}{2}\hat{\gamma}\alpha^2\|\nabla \mathcal{S}(\x_k)\|^2 . \\
\end{aligned}
\end{equation}
Taking expectations on both sides, we can obtain
\begin{equation}
\begin{aligned}
\mathbb{E}&[V(\Theta_{k+1})-V(\Theta_k)] \\
&\leq \mathbb{E}[-\alpha\nabla V(\Theta_k)^T\nabla \mathcal{S}(\Theta_k) +\frac{1}{2}\hat{\gamma}\alpha^2\|\nabla \mathcal{S}(\Theta_k)\|^2] .
\end{aligned}
\end{equation}
While $V(\Theta_k)$ is deterministic, $V(\Theta_{k+1})$ can be considered to be stochastic due to the random sampling aspect. Therefore, we have
\begin{equation}
\begin{aligned}
\mathbb{E}&[V(\Theta_{k+1})]-V(\Theta_k) \\
&\leq -\alpha\nabla V(\Theta_k)^T\mathbb{E}[\nabla \mathcal{S}(\Theta_k)]+\frac{1}{2}\hat{\gamma}\alpha^2\mathbb{E}[\|\nabla \mathcal{S}(\Theta_k)\|^2] ,
\end{aligned}
\end{equation}
which completes the proof.
\end{proof}
\noindent\textbf{Lemma~\ref{lemma2}}:
Let Assumptions~\ref{assump_1},~\ref{assump_2}, and~\ref{assump_3} hold. The iterates of i-CDSGD (Algorithm~\ref{GCDSGD}) satisfy the following inequality $\forall k\in \mathbb{N}$:
\begin{align}\label{lem2}
\E&[V(\Theta_{k+1})]-V(\Theta_k) \\
&\leq -(r_1-\frac{\hat{\gamma}}{2}\alpha B_m)\alpha\|\nabla V(\Theta_k)\|^2
                       +\frac{\hat{\gamma}}{2}\alpha^2B .
\end{align}
\begin{proof}
Recalling Lemma~\ref{lemma1} and using Assumption~\ref{assump_2} and Remark~\ref{rem1}, we have
\begin{equation}
\begin{aligned}
&\E[V(\Theta_{k+1})]-V(\Theta_k)\leq -r_1\alpha\|\nabla V(\Theta_k)\|^2\\
&+\frac{\hat{\gamma}}{2}\alpha^2\E[\|\nabla \mathcal{S}(\Theta_k)\|^2]\leq -r_1\alpha\|\nabla V(\Theta_k)\|^2\\
&+\frac{\hat{\gamma}}{2}\alpha^2(B+B_m\|\nabla V(\Theta_k)\|^2)\\
&=-(r_1-\frac{\hat{\gamma}}{2}\alpha B_m)\alpha\|\nabla V(\Theta_k)\|^2+\frac{\hat{\gamma}}{2}\alpha^2B
\end{aligned}
\end{equation}
which completes the proof.
\end{proof}
\noindent\textbf{Proposition~\ref{prop1}}:
Let Assumptions~\ref{assump_1},~\ref{assump_2},~\ref{assump_4} hold. The iterates of g-CDSGD (Algorithm~\ref{GCDSGD}) satisfy the following inequality $\forall k\in \mathbb{N}$, when $\alpha$ satisfies Eq.~\ref{step_size},
\begin{equation}
\E[\|\theta^j_k - s_k\|]\leq \frac{\omega\alpha h}{1-\hat{\lambda}_2} ,
\end{equation}
where $s_k = \frac{1}{N}\sum_{j=1}^{N}\theta_k^j$, $\hat{\lambda}_2$ is the second-largest eigenvalue of the matrix $\mathbf{Q}=(1-\omega)(\Pi\otimes I_d)+\omega I_{Nd}$.
\begin{proof}
Rewriting the expression~\ref{gcd-sgd} in another form yields $\Theta_{k+1}=\mathbf{Q}\Theta_k - \omega\alpha\mathbf{g}(\Theta_k)$. Recursively applying the new form of Eq.~\ref{gcd-sgd} results in the following expression
\begin{equation}
\Theta_k = -\omega\alpha\sum_{o=0}^{k-1}\mathbf{Q}^{k-1-o}\mathbf{g}(\Theta_k)
\end{equation}
which follows from that the initial value of $\Theta_k$ is set 0.
Let $\mathbf{s}_k = [s_k;s_k;...;s_k]\in\mathbb{R}^{Nd}$ such that $\mathbf{s}_k=\frac{1}{Nd}(\mathbf{1}_{Nd}\mathbf{1}_{Nd}^T)\Theta_k$. Therefore, we have
\begin{equation}
\begin{aligned}
&\|\theta_k^j-s_k\|\leq\|\Theta_k-\mathbf{s}_k\|\\
&=\|\Theta_k-\frac{1}{Nd}(\mathbf{1}_{Nd}\mathbf{1}_{Nd}^T)\Theta_k\|\\
&=\|-\omega\alpha\sum_{o=0}^{k-1}\mathbf{Q}^{k-1-o}\mathbf{g}(\Theta_o)\\
&~~~~~+\omega\alpha\sum_{o=0}^{k-1}\frac{1}{Nd}(\mathbf{1}_{Nd}\mathbf{1}_{Nd}^T\mathbf{Q}^{k-1-o})\mathbf{g}(\Theta_o)\|\\
&=\|-\omega\alpha\sum_{o=0}^{k-1}\mathbf{Q}^{k-1-o}\mathbf{g}(\Theta_o)\\
&~~~~~+\omega\alpha\sum_{o=0}^{k-1}\frac{1}{Nd}(\mathbf{1}_{Nd}\mathbf{1}_{Nd}^T)\mathbf{g}(\Theta_k)\|\\
&=\omega\alpha\|\sum_{o=0}^{k-1}(\mathbf{Q}^{k-1-o}-\frac{1}{Nd}\mathbf{1}_{Nd}\mathbf{1}_{Nd}^T)\mathbf{g}(\Theta_o)\|\\
&\leq\omega\alpha\sum_{o=0}^{k-1}\|\mathbf{Q}^{k-1-o}-\frac{1}{Nd}\mathbf{1}_{Nd}\mathbf{1}_{Nd}^T\|\|\mathbf{g}(\Theta_o)\|\\
&=\omega\alpha\sum_{o=0}^{k-1}\hat{\lambda}_2^{k-1-o}\|\mathbf{g}(\Theta_o)\|,
\end{aligned}
\end{equation}
where the third equality follows from that $\frac{1}{Nd}\mathbf{1}_{Nd}\mathbf{1}_{Nd}^T\mathbf{Q}=\frac{1}{Nd}\mathbf{1}_{Nd}\mathbf{1}_{Nd}^T$, the second inequality is obtained by using Cauchy-Schwartz inequality, $\hat{\lambda}_2\textless 1$. 

Therefore, the following relationships can be obtained:
\begin{equation}
\begin{aligned}
\mathbb{E}[\|\theta_k^j-s_k\|]
&\leq\omega\alpha\mathbb{E}[\sum_{o=0}^{k-1}\hat{\lambda}_2^{k-1-o}\|\mathbf{g}(\Theta_o)\|]\\
&=\omega\alpha\sum_{o=0}^{k-1}\hat{\lambda}_2^{k-1-o}\mathbb{E}[\|\mathbf{g}(\Theta_o)\|]\\
&\leq\frac{\omega\alpha h}{1-\hat{\lambda}_2} ,
\end{aligned}
\end{equation}
which completes the proof.
\end{proof}
Similarly, the consensus bound for i-CDSGD is shown as follows.

\noindent\textbf{Proposition~\ref{prop2}}:
Let Assumptions~\ref{assump_1},~\ref{assump_2},~\ref{assump_4} hold. The iterates of i-CDSGD (Algorithm~\ref{ICDSGD}) satisfy the following inequality $\forall k\in \mathbb{N}$, when $\alpha$ satisfies $0\textless\alpha\leq\frac{r_1-(1-\lambda_N^\tau)B_m}{\gamma_mB_m}$:
\begin{equation}
\E[\|\theta^j_k - s_k\|]\leq \frac{\alpha h}{1-\lambda_2^\tau}
\end{equation}
where $s_k = \frac{1}{N}\sum_{j=1}^{N}\theta_k^j$.
\begin{proof}
Rewriting Equation~\eqref{con_update_icdsgd} yields 
$$\Theta_{k+1}=\mathbf{P}^\tau\Theta_k - \alpha\mathbf{g}(\Theta_k).$$ 
Recursively applying the new form of Equation~\eqref{con_update_icdsgd} results in the following expression:
\begin{equation}
\Theta_k = -\alpha\sum_{o=0}^{k-1}\mathbf{P}^{\tau(k-1-o)}\mathbf{g}(\Theta_k)
\end{equation}
which follows from the fact that that the initial value of $\Theta_k$ is set 0.

Let $\mathbf{s}_k = [s_k;s_k;...;s_k]\in\mathbb{R}^{Nd}$ such that 
$$\mathbf{s}_k=\frac{1}{Nd}(\mathbf{1}_{Nd}\mathbf{1}_{Nd}^T)\Theta_k.$$ 
Therefore, we have:
\begin{equation}
\begin{aligned}
\|\theta_k^j-s_k\|&\leq\|\Theta_k-\mathbf{s}_k\|\\
&=\|\Theta_k-\frac{1}{Nd}(\mathbf{1}_{Nd}\mathbf{1}_{Nd}^T)\Theta_k\|\\
&=\|-\alpha\sum_{o=0}^{k-1}\mathbf{P}^{\tau(k-1-o)}\mathbf{g}(\Theta_o)\\
&~~~~~+\alpha\sum_{o=0}^{k-1}\frac{1}{Nd}(\mathbf{1}_{Nd}\mathbf{1}_{Nd}^T\mathbf{P}^{\tau(k-1-o)})\mathbf{g}(\Theta_o)\|\\
&=\|-\alpha\sum_{o=0}^{k-1}\mathbf{P}^{\tau(k-1-o)}\mathbf{g}(\Theta_o)\\
&~~~~~+\alpha\sum_{o=0}^{k-1}\frac{1}{Nd}(\mathbf{1}_{Nd}\mathbf{1}_{Nd}^T)\mathbf{g}(\Theta_k)\|\\
&=\alpha\|\sum_{o=0}^{k-1}(\mathbf{P}^{\tau(k-1-o)}-\frac{1}{Nd}\mathbf{1}_{Nd}\mathbf{1}_{Nd}^T)\mathbf{g}(\Theta_o)\|\\
&\leq\alpha\sum_{o=0}^{k-1}\|\mathbf{P}^{\tau(k-1-o)}-\frac{1}{Nd}\mathbf{1}_{Nd}\mathbf{1}_{Nd}^T\|\|\mathbf{g}(\Theta_o)\|\\
&=\alpha\sum_{o=0}^{k-1}\lambda_2^{\tau(k-1-o)}\|\mathbf{g}(\Theta_o)\|,
\end{aligned}
\end{equation}
where the third equality follows from that $\frac{1}{N}\mathbf{1}_{N}\mathbf{1}_{N}^T\mathbf{P}=\frac{1}{N}\mathbf{1}_{N}\mathbf{1}_{N}^T$, and the second inequality is obtained by using the Cauchy-Schwartz inequality. Therefore, 
\begin{equation}
\begin{aligned}
\mathbb{E}[\|\theta_k^j-s_k\|]&\leq\alpha\mathbb{E}[\sum_{o=0}^{k-1}\lambda_2^{\tau(k-1-o)}\|\mathbf{g}(\Theta_o)\|]\\
&=\alpha\sum_{o=0}^{k-1}\lambda_2^{\tau(k-1-o)}\mathbb{E}[\|\mathbf{g}(\Theta_o)\|]\\
&\leq\frac{\alpha h}{1-\lambda_2^\tau} .
\end{aligned}
\end{equation}
which completes the proof.
\end{proof}

We next show the main results for g-CDSGD and i-CDSGD with both strongly convex and nonconvex objective functions. 

\noindent\textbf{Theorem~\ref{thm1}}~{(Convergence of g-CDSGD to global optimum, strongly convex case)}:
Let Assumptions~\ref{assump_1},~\ref{assump_2} and~\ref{assump_3} hold. When the step size satisfies Eq.~\ref{step_size}, the iterates of g-CDSGD (Algorithm~\ref{GCDSGD}) satisfy the following inequality $\forall k\in \mathbb{N}$:
\begin{equation}
\E[D_k]\leq C_1^{k-1}D_1+C_2\sum_{q=0}^{k-1}C_1^{q}
\end{equation}
where $D_k =V(\Theta_k)-V^*, C_1 = 1-(\omega\alpha H_m+\frac{1-\omega}{2}(1-\lambda_2))r_1, C_2 = \frac{(\alpha^2\gamma_m\omega+\alpha(1-\omega)(1-\lambda_N))B}{2}$.
\begin{proof}
Recalling Lemma~\ref{lemma2} and using Definition~\ref{strong_convexity_smooth} yield:
\begin{equation}
\begin{aligned}
\E[V&(\Theta_{k+1})]-V(\Theta_k) \\
& \leq -(r_1-\frac{\hat{\gamma}}{2}\alpha B_m)\alpha\|\nabla V(\Theta_k)\|^2\\
&+\frac{\hat{\gamma}}{2}\alpha^2B\leq -\frac{1}{2}\alpha r_1\|\nabla(\Theta_k)\|^2+\frac{\alpha^2\hat{\gamma}B}{2}\\
&\leq -\alpha r_1\hat{H}(V(\Theta_k)-V^*)+\frac{\alpha^2\hat{\gamma}B}{2} .
\end{aligned}
\end{equation}
The second inequality follows from that $\alpha\leq \frac{r_1}{\hat{\gamma}B_m}$, which is implied by Eq.~\ref{step_size}. The expectation taken in the above inequalities is only related to $\theta_{k+1}$. Hence, recursively taking the expectation and subtracting $V^*$ from both sides, we get:
\begin{equation}
\E[V(\Theta_{k+1})-V^*]\leq (1-\alpha\hat{H}r_1)\E[V(\Theta_k)-V^*]+\frac{\alpha^2\hat{\gamma}B}{2} .
\end{equation}
As $0\textless \alpha\hat{H}r_1\leq\frac{\hat{H}r_1^2}{\hat{\gamma}B_m}\leq \frac{\hat{H}r_1^2}{\hat{\gamma}r_1^2}=\frac{\hat{H}}{\hat{\gamma}}\leq 1$, the conclusion follows by applying Eq.~\ref{thm1} recursively through iteration $k\in \N$ and letting $D_k =V(\Theta_k)-V^*, C_1 = 1-(\omega\alpha H_m+\frac{1-\omega}{2}(1-\lambda_2))r_1, C_2 = \frac{(\alpha^2\gamma_m\omega+\alpha(1-\omega)(1-\lambda_N))B}{2}$.
\end{proof}
\noindent\textbf{Theorem~\ref{thm2}}~{(Convergence  of i-CDSGD to global optimum, strongly convex case)}
Let Assumptions~\ref{assump_1},~\ref{assump_2} and~\ref{assump_3} hold. When the step size satisfies $0\textless\alpha\leq\frac{r_1-(1-\lambda_N^\tau)B_m}{\gamma_mB_m}$, the iterates of i-CDSGD (Algorithm~\ref{ICDSGD}) satisfy the following inequality $\forall k\in \mathbb{N}$:
\begin{equation}
\E[D_k]\leq C_3^{k-1}D_1+C_4\sum_{q=0}^{k-1}C_3^{q}
\end{equation}
where $D_k =V(\Theta_k)-V^*, C_3 = 1-(\alpha H_m+\frac{1}{2}(1-\lambda_2^\tau))r_1, C_4 = \frac{(\alpha^2\gamma_m+\alpha(1-\lambda_N^\tau))B}{2}$.
\begin{proof}
We omit the proof here and one can easily get it following the proof techniques shown for Theorem~\ref{thm1}. The desired result is obtained by replacing $C_1$ with $C_3$ and $C_2$ with $C_4$, respectively. 
\end{proof}
\noindent\textbf{Theorem~\ref{thm3}}~{(Convergence of g-CDSGD to a first-order stationary point, non-convex case)}:
Let Assumptions~\ref{assump_1},~\ref{assump_2}, and~\ref{assump_3} hold. When the step size satisfies Eq.~\ref{step_size}, the iterates of g-CDSGD (Algorithm~\ref{GCDSGD}) satisfy the following inequality $\forall K\in \mathbb{N}$:
\begin{equation}\label{part1_thm2}
\begin{aligned}
\E[&\frac{1}{K}\sum_{k=1}^{K}\|\nabla V(\Theta_k)\|^2] \\
&\leq\frac{(\omega\gamma_m\alpha+(1-\omega)(1-\lambda_N))B}{r_1}+\frac{2(V(\Theta_1)-V_{\text{inf}})}{Kr_1\alpha}
\end{aligned}
\end{equation}
\begin{proof}
Recalling Lemma~\ref{lemma2} and taking expectations on both sides lead to the following relation:
\begin{equation}\label{nonconvex_descent}
\begin{aligned}
\E[V(&\Theta_{k+1})]-\E[V(\Theta_k)]\\
&\leq -(r_1-\frac{\hat{\gamma}\alpha B_m}{2})\alpha\E[\|\nabla V(\Theta_k)\|^2]+\frac{\hat{\gamma}\alpha^2B}{2}.
\end{aligned}
\end{equation}
If the step size is such that $\alpha\leq\frac{r_1}{\hat{\gamma}B_m}$, we get:
\begin{equation}
\E[V(\Theta_{k+1})]-\E[V(\Theta_k)]\leq  -\frac{r_1\alpha}{2}\E[\|\nabla V(\Theta_k)\|^2]+\frac{\alpha^2\hat{\gamma}B}{2}.
\end{equation}
Applying the above inequality from 1 to $K$ and summing them up can give the following relation
\begin{equation}
\begin{aligned}
V_{\text{inf}} - V(\Theta_1)&\leq \E[V(\Theta_{k+1})]-V(\Theta_1)\\
&\leq -\frac{r_1\alpha}{2}\sum_{k=1}^{m}\E[\|\nabla V(\Theta_k)\|^2]+\frac{m\alpha^2\hat{\gamma}B}{2} .
\end{aligned}
\end{equation}
The last inequality follows from the Assumption~\ref{assump_3}. Rearrangement of the above inequality, substituting $\hat{\gamma} = \omega\gamma_m+\alpha^{-1}(1-\omega)(1-\lambda_N)$, and dividing by $K$ yields the desired result.
\end{proof}

\noindent\textbf{Theorem~\ref{thm4}}~{(Convergence of i-CDSGD to a first-order stationary point, non-convex case)}:
Let Assumptions~\ref{assump_1},~\ref{assump_2}, and~\ref{assump_3} hold. When the step size satisfies $0\textless\alpha\leq\frac{r_1-(1-\lambda_N^\tau)B_m}{\gamma_mB_m}$, the iterates of i-CDSGD (Algorithm~\ref{ICDSGD}) satisfy the following inequality $\forall K\in \mathbb{N}$:
\begin{equation}\label{part1_thm2}
\begin{aligned}
\E[&\frac{1}{K}\sum_{k=1}^{K}\|\nabla V(\Theta_k)\|^2] \\
&\leq\frac{(\gamma_m\alpha+(1-\lambda_N^\tau))B}{r_1}+\frac{2(V(\Theta_1)-V_{\text{inf}})}{Kr_1\alpha}
\end{aligned}
\end{equation}
\begin{proof}
The proof for this theorem is rather similar to the one provided for Theorem~\ref{thm3} above, and we omit the details. 
\end{proof}

We now give the proof for g-CDMSGD as well as the proof for i-CDMSGD for completeness.  Note that for i-CDMSGD, the strongly convex and smooth constants are different from those of g-CDMSGD. 
We first present several auxiliary technical lemmas.

Define
\[v_k = \textnormal{arg}\min\limits_{\Theta\in\mathbb{R}^{Nd}}\phi_k(\Theta)\]
\begin{lem}\label{lemma3}
The process generated by the Eq.~\ref{estimate} preserves the canonical form of functions $\{\phi_k(\Theta)\}$ when $\phi_1(\Theta) = \phi^*_1 + \frac{\hat{H}}{2}\|\Theta - \Theta_1\|^2$:
\begin{equation}
\phi_k(\Theta) = \phi^*_k + \frac{\hat{H}}{2}\|\Theta - \Theta_k\|^2
\end{equation}
\end{lem}
\begin{lem}\label{lemma4}
If $\alpha\leq\textnormal{min}\{\frac{r_1-(1-\omega)(1-\lambda_N)B_m}{\omega B_m\gamma_m}, \frac{1}{\hat{H}}\}$, then 
the sequences $\{v_k\}$ and $\{v_k-\mathbf{y}_k\}$ are defined as follows:
\begin{subequations}\label{sequence}
\begin{alignat}{2}
  &v_{k+1} = (1-\sqrt{\hat{H}\alpha})v_k + \sqrt{\hat{H}\alpha}\mathbf{y}_k - \sqrt{\frac{\alpha}{\hat{H}}}\mathcal{S}(\mathbf{y}_k)\\
  &v_k - \mathbf{y}_k = \frac{1}{\sqrt{\hat{H}\alpha}}(\mathbf{y}_k - \Theta_k)
\end{alignat}
\end{subequations}
\end{lem}
The proof of both Lemmas follow from~\cite{nesterov2013introductory}. We also have:
\begin{lem}\label{lemma5}
Let Assumptions~\ref{assump_1},~\ref{assump_2},~\ref{assump_3}, and~\ref{assump_4} hold. If $\alpha\leq\textnormal{min}\{\frac{r_1-(1-\omega)(1-\lambda_N)B_m}{\omega B_m\gamma_m}, \frac{1}{\hat{H}}, \frac{1}{2\hat{\gamma}}\}$, then for $\forall k\in\mathbb{N}$, we have:
\begin{equation}\label{estimate_1}
\mathbb{E}[\phi_k(\Theta)]\leq V(\Theta)+(1-\sqrt{\hat{H}\alpha})^{k-1}(\phi_1(\Theta) - V(\Theta)),
\end{equation}
\begin{equation}\label{estimate_2}
\begin{aligned}
&\mathbb{E}[V(\Theta_k)]\leq\mathbb{E}\Bigg[\phi^*_k+\sum_{p=1}^{k-1}(1-\sqrt{\hat{H}\alpha})^{k-1-p}\\
&\bigg\{-\frac{\hat{H}}{2}\frac{1-\sqrt{\hat{H}\alpha}}{\sqrt{\hat{H}\alpha}}\|\Theta_p-\mathbf{y}_p\|^2+\alpha\|\nabla V(\mathbf{y}_p)-\mathcal{S}(\mathbf{y}_p)\|^2\bigg\}\Bigg]
\end{aligned}
\end{equation}
\end{lem}
The proof of this lemma follows from Lemmas~\ref{lemma3} and~\ref{lemma4}, Lemma 1 of~\cite{nitanda2014stochastic}, and the expressions:
\begin{align*}
&(\nabla V(\mathbf{y}_k),\mathcal{S}(\mathbf{y}_k)) \\
&= \frac{1}{2}(\|\nabla V(\mathbf{y}_k)\|^2 +\|\mathcal{S}(\mathbf{y}_k)\|^2 - \|\nabla V(\mathbf{y}_k)-\mathcal{S}(\mathbf{y}_k)\|^2) ,
\end{align*}
\begin{align*}
\|\mathcal{S}(\mathbf{y}_k)\|^2 &\leq2(\|\nabla V(\mathbf{y}_k)\|^2 + \|\nabla V(\mathbf{y}_k)-\mathcal{S}(\mathbf{y}_k)\|^2), \\
\|\nabla V(\mathbf{y}_k)\|^2 &\leq 2(\|\mathcal{S}(\mathbf{y}_k)\|^2 + \|\nabla V(\mathbf{y}_k)-\mathcal{S}(\mathbf{y}_k)\|^2).
\end{align*}
The last two inequalities directly follow from the triangle inequality.

\noindent\textbf{Theorem~\ref{thm5}} {(Convergence of g-CDMSGD, strongly convex case)}:
Let Assumptions~\ref{assump_1},~\ref{assump_2},~\ref{assump_3}, and~\ref{assump_4} hold. If the step size $\alpha$ satisfies $\alpha\leq\textnormal{min}\{\frac{r_1-(1-\omega)(1-\lambda_N)B_m}{\omega B_m\gamma_m}, \frac{1}{\hat{H}}, \frac{1}{2\hat{\gamma}}\}$, then we have
\begin{equation}
\begin{aligned}
&\mathbb{E}[V(\Theta_k) - V^*]\leq (1-\sqrt{\hat{H}\alpha})^{k-1}(\phi_1^*-V^*)\\
&+\sqrt{\frac{\alpha}{\hat{H}}}(B+B_{V}G^2)
\end{aligned}
\end{equation}
\begin{proof}
From Lemma~\ref{lemma5}, it can be obtained that
\begin{equation}\label{lemma5_1}
\begin{aligned}
\mathbb{E}[V(\Theta_k)]&\leq\mathbb{E}\Bigg[\phi^*_k+\sum_{p=1}^{k-1}(1-\sqrt{\hat{H}\alpha})^{k-1-p}\\
                       &\bigg\{\alpha\|\mathcal{S}(\mathbf{y}_k)-\nabla V(\mathbf{y}_k)\|^2\bigg\}\Bigg]
\end{aligned}
\end{equation}
The last inequality follows from that the coefficient $-\frac{\hat{H}}{2}\frac{1-\sqrt{\hat{H}\alpha}}{\sqrt{\hat{H}\alpha}}\textless 0$. Recalling Eq.~\ref{estimate_1} and letting $\Theta = \Theta^*$, and combining Eq.~\ref{lemma5_1}, we have
\begin{equation}
\begin{aligned}
\mathbb{E}[V(\Theta_k)]&\leq\mathbb{E}[V^*+(1-\sqrt{\hat{H}\alpha})^{k-1}(\phi^*_1-V^*)]\\
                       &+\mathbb{E}\Bigg[\sum_{p=1}^{k-1}(1-\sqrt{\hat{H}\alpha})^{k-1-p}\\
                       &\bigg\{\alpha\|\mathcal{S}(\mathbf{y}_k)-\nabla V(\mathbf{y}_k)\|^2\bigg\}\Bigg]
\end{aligned}
\end{equation}
As $\mathbb{E}[\|\mathcal{S}(\mathbf{y}_k) - \nabla V(\mathbf{y}_k)\|^2]\leq B+B_{V}G^2$, therefore, the following inequality can be acquired
\begin{equation}
\begin{aligned}
\mathbb{E}[V(\Theta_k)-V^*]&\leq(1-\sqrt{\hat{H}\alpha})^{k-1}(\phi^*_1-V^*)\\
                           &+\mathbb{E}\Bigg[\sum_{p=1}^{k-1}(1-\sqrt{\hat{H}\alpha})^{k-1-p}(B+B_{V}G^2)\Bigg]
\end{aligned}
\end{equation}
Using $\sum_{p=1}^{k-1}(1-\sqrt{\hat{H}\alpha})^{k-1-p}\leq\sum_{t=0}^{\infty}(1-\sqrt{\hat{H}\alpha})^t=\frac{1}{\sqrt{\hat{H}\alpha}}$ completes the proof.
\end{proof}

\subsection{Discussion on consensus and optimality trade-offs for various algorithms}\label{A2}
As shown in Figure~\ref{Figure1}, we formally denote the consensus bound after sufficient iterations by $d_1$.
Observe that the consensus (upper) bound is a function of the spectral properties of the underlying communication topology (specifically, proportional to $1-\lambda_2$ for g-CDSGD, or $1-\lambda_2^\tau$ for i-CDSGD). Let us consider two illustrative example communication topologies: dense ($\lambda_2 = 0.01$) and sparse ($\lambda_2 = 0.8$). We can observe that with even $\tau = 2$, i-CDSGD has a much smaller consensus bound compared to that of CDSGD for the sparse topology. However, the improvement is negligible for the dense topology. Therefore, in practice one can achieve better consensus with higher $\tau$ for sparser topologies.  
For g-CDSGD, as $d_1$ is also a function of the parameter $\omega$, it can be seen that with an appropriately chosen $\omega$, one can reduce the consensus bound significantly. However, the tuning of $\omega$ can affect the optimality as we discuss later in the paper. For i-CDSGD, the smallest consensus bound is $\alpha h$ when $\tau\to \infty$, which leads to a large communication cost. Considering $\frac{\omega\alpha h}{1-\hat{\lambda}_2}\leq\frac{\alpha h}{1-\lambda_2^\tau}$, we obtain the condition $\omega\leq \frac{1-\lambda_2}{2-\lambda_2-\lambda_2^\tau}$ that guarantees g-CDSGD to have a better consensus bound than i-CDSGD.

In sparse networks, i-CDSGD performs empirically better than CDSGD in terms of optimality; here we attempt to explain why our theory suggests this is the case. For completeness we also compare g-CDSGD with i-CDSGD and CDSGD. 

\textbf{Comparisons between i-CDSGD and g-CDSGD.} We provide optimality bounds (which can be interpreted as the Euclidean distance between $\hat{\theta}$ and $\theta^*$ in Figure~\ref{Figure1}. In this context, we give the upper bound for i-CDSGD, which is 
\[\lim_{k\to\infty}\mathbb{E}[V(\Theta_k)-V*]\leq\frac{B(\alpha\gamma_m+1-\lambda_N^\tau)}{2r_1(H_m+\alpha^{-1}(1-\lambda_2^\tau))},\] 
which demonstrates that the optimality bound is related to $\tau$. Theorem~\ref{thm1} shows the optimality bound of g-CDSGD is a function of $\omega$. We discuss the comparison for the strongly convex case; the non-convex case follows from the similar analysis techniques to obtain the conclusion.
Suppose the following condition holds:
\begin{equation}
\begin{aligned}
&\frac{B[\omega\alpha\gamma_m+(1-\omega)(1-\lambda_N)]}{2r_1(\omega H_m+\alpha^{-1}(1-\omega)(1-\lambda_2))} \\
& \leq\frac{B[\alpha\gamma_m+1-\lambda_N^\tau]}{2r_1(H_m+\alpha^{-1}(1-\lambda_2^\tau))}
\end{aligned}
\end{equation}
which leads to 
\[
\omega\geq\frac{2H_m a-b\gamma_m+(be-da)\alpha^{-1}}{2H_m(a+e)+(ad-be)\alpha^{-1}-\gamma_m(b+d)}
\]
where $a = 1-\lambda_N, b = 1-\lambda_2, e =1-\lambda_N^\tau, d=1-\lambda_2^\tau$.
Let 
\begin{align*}
A_1 &= 2H_m a-b\gamma_m+(be-da)\alpha^{-1}, \\
A_2 &= 2H_m(a+e)+(ad-be)\alpha^{-1}-\gamma_m(b+d).
\end{align*}
To guarantee the lower bound is positive and less than 1, the following condition should be satisfied:
\begin{equation}
A_1\textgreater 0,~A_2\textgreater 0,~A_1\textless A_2 .
\end{equation}
Based on the above condition, we obtain:
\[
c\textless \textnormal{min}\big\{2\frac{2a+e}{2b+d},\frac{2H_me-2(bc-ad)\alpha^{-1}}{dH_m}\big\}
\]
Thus the lower bound for $\omega$ is obtained for the guarantee that g-CDSGD has a better optimal bound than i-CDSGD in strongly convex case. 

\textbf{Comparison between CDSGD and g-CDSGD}. Given the optimality upper bound of CDSGD when $k\to\infty$~\cite{jiang2017collaborative} as follows:
\[\lim_{k\to\infty}\mathbb{E}[V(\Theta_k)-V*]\leq\frac{B(\alpha\gamma_m+1-\lambda_N)}{2r_1(H_m+\alpha^{-1}(1-\lambda_2))},\] we have:
\begin{equation}
\begin{aligned}
&\frac{B[\omega\alpha\gamma_m+(1-\omega)(1-\lambda_N)]}{2r_1(\omega H_m+\alpha^{-1}(1-\omega)(1-\lambda_2))} \\
& \leq\frac{B[\alpha\gamma_m+1-\lambda_N]}{2r_1(H_m+\alpha^{-1}(1-\lambda_2))}
\end{aligned}
\end{equation}
After some mathematical manipulations, we can obtain the following lower bound for $\omega$:
\[\omega\geq \frac{1}{2}\]. 
Combining the lower bound for $\omega$ after comparing i-CDSGD with g-CDSGD, it can be obtained that
\[
\omega\geq\max\bigg\{\frac{1}{2},\frac{2H_m a-b\gamma_m+(be-da)\alpha^{-1}}{2H_m(a+e)+(ad-be)\alpha^{-1}-\gamma_m(b+d)}\bigg\}
\]
Such a result may improve the lower bound for $\omega$ to be tighter. However, since for sparse networks, i-CDSGD outperforms CDSGD, the lower bound for $\omega$ we have shown in the main contents is an enough guarantee for improving the optimality.

\subsection{Additional pseudo-codes of the algorithms}\label{A3}
Applying momentum term to optimization methods has been done for speeding up the convergence rate and it provably shows the efficiency improvement in many problems. This section presents the Nesterov momentum variants of i-CDSGD (Algorithm~\ref{ICDMSGD}) and g-CDSGD (Algorithm~\ref{GCDMSGD}). 

\begin{algorithm}[h]
    \caption{i-CDMSGD}
    \begin{algorithmic}[1]\label{ICDMSGD}
    \STATE \textbf{Initialization}: $\theta_0^j, v_0^j, \;j=1,2,...,N$, $\alpha$, $N$, $\tau$, $m$, $\Pi$, $\mu$
    \STATE Distribute the \textit{Non-IID} training data set to $N$ agents
    \FOR{\textit{each agent}}
      \STATE {Randomly shuffle each data subset}
      \FOR{$k=0:m$}
        \STATE {$t=0$}
        \FOR{$j = 1,...,N$}
          \STATE {$\theta^j_t = \theta^j_k$}
          \STATE {$v^j_t = v^j_k$}
        \ENDFOR
        \FOR{$j = 1,...,N$}
          \WHILE {$t\leq \tau-1$}
            \STATE {$\theta^j_{t+1} =\sum_{l\in Nb(j)}\pi_{jl}\theta^l_t$}
            \STATE {$v^j_{t+1} =\sum_{l\in Nb(j)}\pi_{jl}v^l_t$\COMMENT{\textbf{Momentum Consensus}}}
            \STATE {$t = t+1$}
          \ENDWHILE
         \ENDFOR
        \STATE {$\hat{\theta} = \theta^j_t$}
        \STATE {$\hat{v} = v^j_t$}
        \STATE {$v_{k+1}^j = \hat{\theta} - \theta^j_k + \mu \hat{v} - \alpha g_j(\theta^j_k)$}
        \STATE {$\theta^j_{k+1} = \theta^j_k + v^j_{k+1}$}
      \ENDFOR
    \ENDFOR
    \end{algorithmic}
\end{algorithm}

\begin{algorithm}[h]
    \caption{g-CDMSGD}
    \begin{algorithmic}[1]\label{GCDMSGD}
    \STATE \textbf{Initialization}: $\omega$, $\theta_0^j, v_0^j, \;j=1,2,...,N$, $\alpha$, $N$, $m$, $\Pi$, $\mu$
    \STATE Distribute the \textit{Non-IID} training data set to $N$ agents
    \FOR{\textit{each agent}}
      \STATE {Randomly shuffle each data subset}
      \FOR{$k=0:m$}
        \STATE {$\hat{\theta} =\sum_{l\in Nb(j)}\pi_{jl}\theta^l_t$}
        \STATE {$\hat{v} = \sum_{l\in Nb(j)}\pi_{jl}v^l_t$}
        \STATE {$v_{k+1}^j = (1-\omega)(\hat{\theta} - \theta^j_k + \mu \hat{v}) +\omega\mu v^j_k - \omega\alpha g_j(\theta^j_k + \mu v^j_k)$}
        \STATE {$\theta_{k+1}^j = \theta^j_k + v^j_{k+1}$}
      \ENDFOR
    \ENDFOR
    \end{algorithmic}
\end{algorithm}

\begin{figure*}[!t]
\centering
\begin{subfigure}{.47\textwidth}
  \centering
  \includegraphics[width=\linewidth,trim={0.0cm, 0.0cm, 0.0cm, 0.0cm},clip]{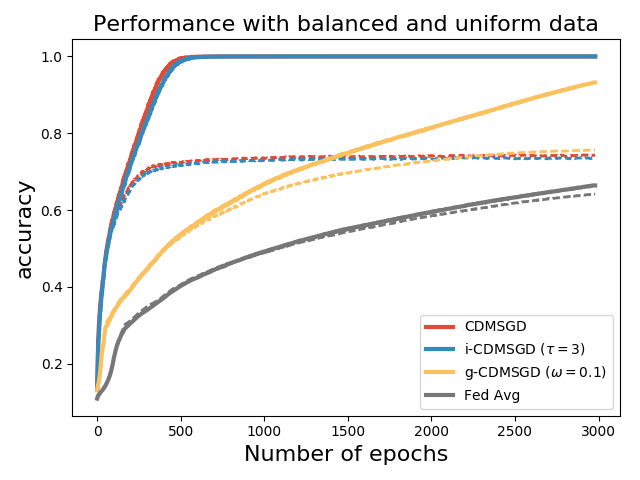}
  \caption{Performance of different algorithms on balanced and uniformly distributed data among agents. (Dashed lines represent test accuracy \& solid lines represent training accuracy.)} 
  \label{Fig:3a}
\end{subfigure}
\hfill
\begin{subfigure}{.47\textwidth}
  \centering
  \includegraphics[width=\linewidth,trim={0.0cm, 0.0cm, 0.0cm, 0.0cm},clip]{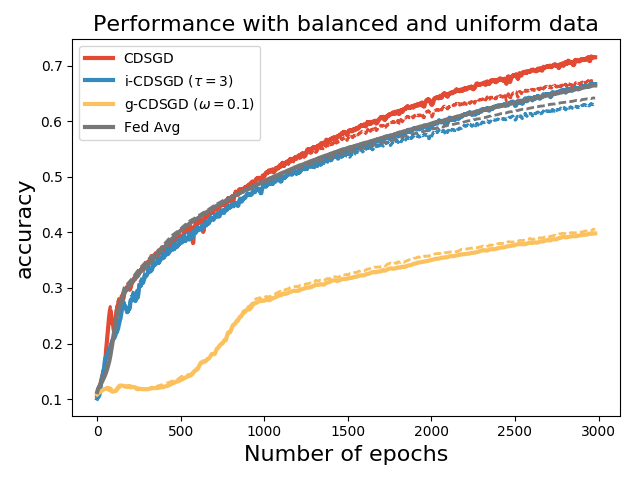}
  \caption{Performance of different non-momentum versions of the algorithms. (Dashed lines represent test accuracy \& solid lines represent training accuracy.)}
  \label{Fig:3b}
\end{subfigure}

\begin{subfigure}{.47\textwidth}
  \centering
  \includegraphics[width=\linewidth,trim={0.0cm, 0.0cm, 0.0cm, 0.0cm},clip]{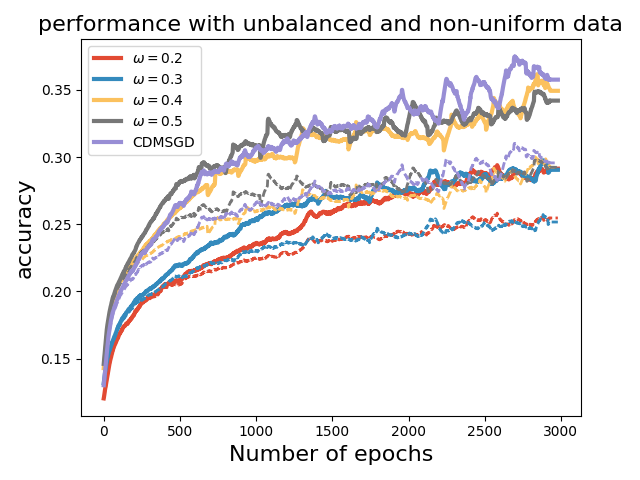}
  \caption{Performance of g-CDMSGD algorithm with different $\omega$ values with an unbalanced and non-uniform distribution of data (20\% non uniformity).}
\label{Fig:3c}
\end{subfigure}%
\hfill
\begin{subfigure}{.47\textwidth}
  \centering
  \includegraphics[width=\linewidth,trim={0.0cm, 0.0cm, 0.0cm, 0.0cm},clip]{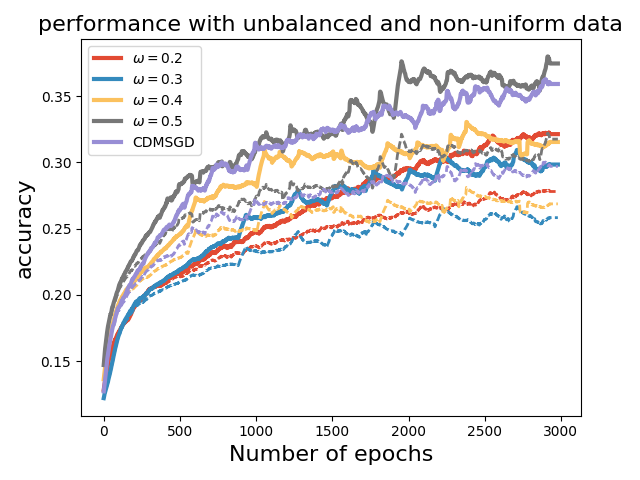}
  \caption{Performance of g-CDMSGD algorithm with different $\omega$ values with an unbalanced and non-uniform distribution of data (40\% non uniformity).}
  \label{Fig:3d}
\end{subfigure}

\begin{subfigure}{.47\textwidth}
  \centering
  \includegraphics[width=\linewidth,trim={0.0cm, 0.0cm, 0.0cm, 0.0cm},clip]{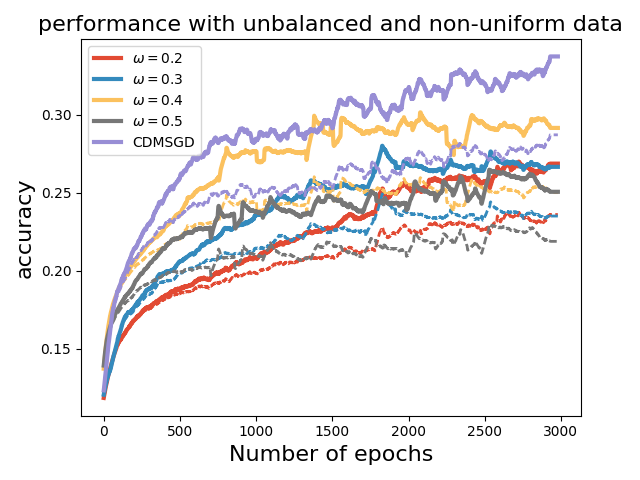}
  \caption{Performance of g-CDMSGD algorithm with different $\omega$ values with an unbalanced and non-uniform distribution of data (60\% non uniformity).}
\label{Fig:3e}
\end{subfigure}%
\hfill
\begin{subfigure}{.47\textwidth}
  \centering
  \includegraphics[width=\linewidth,trim={0.0cm, 0.0cm, 0.0cm, 0.0cm},clip]{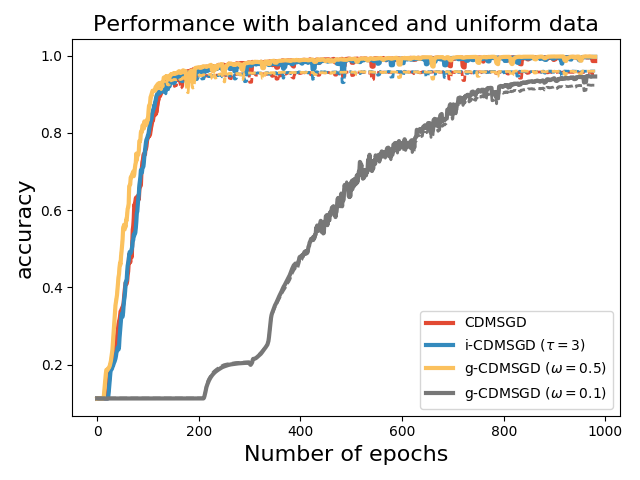}
  \caption{Performance of different algorithms with a balanced data distribution on MNIST dataset}
  \label{Fig:3f}
\end{subfigure}

\caption{Results from numerical experiments on several data distributions for CIFAR10 and also one experiment on MNIST data set}\vspace{-10pt}
\label{Fig:extra_results}
\end{figure*}

\subsection{Additional Experimental Results}\label{A4}
In all our experiments. we consider the number of agents to be 5. We choose the following sparse agent interaction matrix for all our experiments.
$$
\pi =
\begin{bmatrix}
    0.34       & 0.33 & 0.0 & 0. & 0.33 \\
    0.33       & 0.34 & 0.33 & 0.0 & 0.0 \\
    0.0       & 0.33 & 0.34 & 0.33 & 0.0 \\
    0.0       & 0.0 & 0.33 & 0.34 & 0.33 \\
    0.33       & 0.0 & 0.0 & 0.33 & 0.34 \\
\end{bmatrix}$$
More results are shown in Figure.~\ref{Fig:extra_results}. In Figure~\ref{Fig:3a}, we see that fluctuations in the average accuracy are almost negligible for the case where each agent gets balanced and uniformly distributed dataset. Algorithm i-CDMSGD performs as good as CDMSGD. We also notice that g-CDMSGD has a lower convergence rate but achieves slightly better test error which shows similar trend with Unbalanced data distribution case shown in Figure~\ref{Fig:2a}. Figure~\ref{Fig:3b} shows the performance of the non-momentum versions of the same settings. Algorithms i-CDSGD and CDSGD perform similar to Federated Averaging whereas g-CDSGD is slow but the generalization gap is lesser.

For all the experiments until this point, each agent is allocated data from a uniform distribution of data (assured by shuffling of the data). However, it is possible that each agent can have non-uniformity in the distribution of data they are receiving. One of the aspect of non-uniformity is when each agent gets samples biased towards a few (not all) classes and gets very few samples of other classes. Note that this kind of distribution is referred as non-iid data distribution in Federated Averaging\cite{mcmahan2016communication}. For simulating this, we allocate a portion of samples pertaining to a class to a specific agent and the other portion will be pooled, shuffled and distributed. Figure~\ref{Fig:3c}-~\ref{Fig:3e} represents the performance of different algorithms with different percentage of non-uniform distribution of data (percentage of data per class allocated without any shuffling). For Figure~\ref{Fig:3c}, we split 20\% of data pertaining to two classes to an each agent. Thus, each agent has a bias of $\approx30\%$  towards a class. In such a non-uniform distribution of data, the performance of each agent fluctuates a lot more than the other the uniform distribution of data. With several values of $\omega$ we see that as the value of $\omega$ increases, the performance is close to CDMSGD and is even slightly better than it. At the same time, as the percentage of non-uniformity is increased to 60\%, we see that the increasing $\omega$ deteriorates the performance. This can be corroborated with the increase in the agent level difference in the performance and lack of consensus as well as more emphasis on local gradient updates ($\omega = 0.5$). Since, the algorithms have not reached stability, we could not compute the degree of consensus among the agents.

Finally, we also compare our proposed algorithms to CDMSGD on another benchmark dataset - MNIST. The performance of the algorithms is shown in Figure~\ref{Fig:3f} which follows similar trend as observed for CIFAR-10.

\end{document}